\documentclass{article}

\usepackage{microtype}
\usepackage{graphicx}
\usepackage{subcaption}
\usepackage{booktabs} 

\usepackage[preprint]{icml2026}

\usepackage{amsmath}
\usepackage{mathtools}
\usepackage{amsthm}
\usepackage{amsfonts}
\usepackage{xspace}
\usepackage{subcaption}
\usepackage{placeins}
\usepackage{float}
\usepackage{makecell}
\usepackage{booktabs}
\usepackage{pifont}
\usepackage{xcolor}
\usepackage{smartref}
\usepackage{bm}
\usepackage{amssymb}
\usepackage{multirow}
\usepackage{enumitem}

\newcommand{\method}{\textsc{SpatialAlign}\xspace}
\newcommand{\ourmatric}{\textsc{DSR-Score}\xspace}
\newcommand{\ourdataset}{\textsc{DSR-Dataset}\xspace}
\newcommand{\onleft}{\textcolor{teal}{on the $\overset{\leftarrow}{\text{left}}$ of}\xspace}
\newcommand{\ontop}{\textcolor{cyan}{on the $\overset{\uparrow}{\text{top}}$ of}\xspace}
\newcommand{\onright}{\textcolor{magenta}{on the $\overset{\rightarrow}{\text{right}}$ of}\xspace}
\newcommand{\toleft}{\textcolor{brown}{to the $^\leftarrow$left of}\xspace}
\newcommand{\totop}{\textcolor{pink}{to the top$^\uparrow$ of}\xspace}
\newcommand{\toright}{\textcolor{orange}{to the right$^\rightarrow$ of}\xspace}

\usepackage[
    colorlinks=true,
    linkcolor=magenta,
    citecolor=magenta,
    urlcolor=magenta
]{hyperref}

\usepackage[capitalize,noabbrev]{cleveref}

\theoremstyle{plain}

\theoremstyle{definition}

\theoremstyle{remark}

\usepackage[textsize=tiny]{todonotes}

\begin{document}

\twocolumn[{
\icmltitle{\method: Aligning Dynamic Spatial Relationships in Video Generation}
 


  \icmlsetsymbol{equal}{*}

  \begin{icmlauthorlist}
    \icmlauthor{Fengming Liu}{yyy}
    \icmlauthor{Tat-Jen Cham}{yyy}
    \icmlauthor{Chuanxia Zheng}{yyy}
  \end{icmlauthorlist}

  \icmlaffiliation{yyy}{College of Computing and Data Science, Nanyang Technological University, 50 Nanyang Avenue, Singapore 639798}

  \icmlcorrespondingauthor{Tat-Jen Cham}{astjcham@ntu.edu.sg}


  \vskip 0.1in

    \begin{center}
\centering
\includegraphics[width=\linewidth]{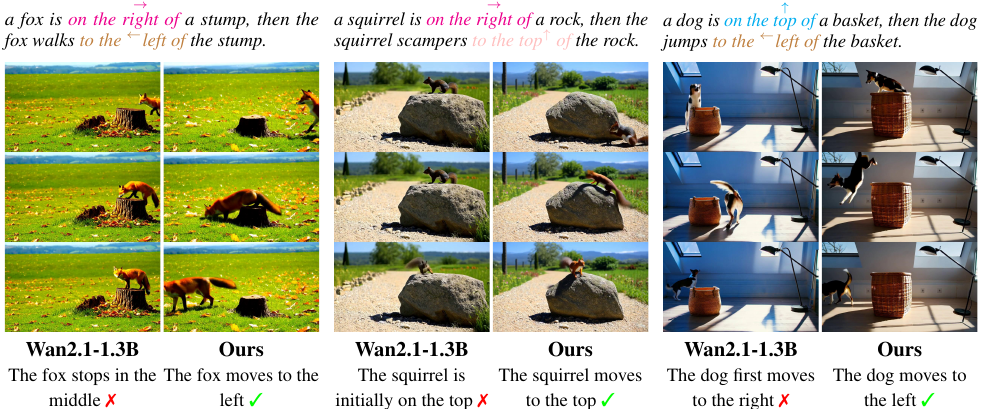}
\captionof{figure}{\textbf{Examples of spatial relationship reasoning.}
Given a prompt specifying the spatial relationships (top),
our fine-tuned model (right) generates videos that correctly reflect the desired change in spatial relationships,
while the baseline model (left) fails to do so.
}
\label{fig:teaser}
\end{center}

    }
]



\printAffiliationsAndNotice{}  

\begin{abstract}

Most text-to-video (T2V) generators prioritize aesthetic quality,
but often ignoring the spatial constraints in the generated videos.
In this work,
we present \method,
a self-improvement framework that enhances T2V models' capabilities to depict \textit{Dynamic Spatial Relationships} (DSR) specified in text prompts.
We present a zeroth-order regularized Direct Preference Optimization (DPO) to fine-tune T2V models towards better alignment with DSR.
Specifically,
we design \ourmatric,
a geometry-based metric that quantitatively measures the alignment between generated videos and the specified DSRs in prompts,
which is a step forward from prior works that rely on VLM for evaluation.
We also conduct a dataset of text-video pairs with diverse DSRs to facilitate the study.
Extensive experiments demonstrate that our fine-tuned model significantly outperforms the baseline in spatial relationships. The code will be released in \href{https://github.com/fengming001ntu/SpatialAlign}{\textit{Link}}.


\end{abstract}
\section{Introduction}
\label{sec:intro}


We live in a spatially and temporally structured world. When things move around and interact with the environment, we as humans often interpret these actions within the context of shared spatial relationships that change over time.
A cat,
for example,
may first be sitting on the top of a table,
and later moves to the right of it.
Humans have a remarkable ability to understand, reason about, and even plan such spatial relationships.
Likewise, developing similar capabilities in AI systems is crucial for a wide range of applications,
like robotics,
and physical world modeling~\cite{chen2024spatialvlm}.

We consider the task of generating plausible videos that accurately reflect \emph{Dynamic Spatial Relationship (DSR)} instructions specified in text prompts.
In particular,
we introduce a simple yet representative task,
where an \emph{animal} moves with respect to a \emph{static object}, leading to a change in the relative spatial relationship
described in a prompt.
For example,
given the leftmost example in~\cref{fig:teaser},
our goal is to generate a video where a fox first appears on the right side of the stump,
and then moves to the left side.
Surprisingly, we find that state-of-the-art T2V models~\cite{wan2025,yang2024cogvideox} often fail to capture such simple DSR instructions reliably.

Many recent contributions like GLIGEN~\cite{li2023gligen} and InstanceDiffusion~\cite{wang2024instancediffusion} have shown promising results in modeling spatial control in generators,
but these methods only work for \emph{static} images
and rely on extra inputs such as bounding boxes (bboxes).
Instead,
we ask \emph{whether
DSRs can be accurately scripted
into generated videos
from text prompts alone}.
In this paper,
we take an initial step towards realizing this goal, going beyond static spatial relationships in images.

Our first contribution is to develop
\textbf{\ourmatric},
a metric to access the correctness of DSR in generated videos.
The latest works like
VBench-2.0~\cite{zheng2025vbench},
3DSRBench~\cite{ma20253dsrbench}
and SpatialBench~\cite{cai2025spatialbot} have proposed to use vision-language models (VLM)~\cite{hurst2024gpt,Qwen2.5-VL} to evaluate such object-object spatial and temporal relationships over time.
However,
we find that VLM-based evaluation is not reliable in this task
(as shown in~\cref{fig:VLM}).
This is partly due to the current VLMs' limited spatial reasoning capabilities~\cite{ma2025spatialreasoner,ma20253dsrbench,batra2025spatialthinker},
especially in dynamic settings.
We thus propose to leverage \emph{geometric} principles to design a more reliable and fine-grained evaluation metric for DSR.
Spatial relationships (SR),
for example ``\emph{on the left of}'' or ``\emph{above}'',
have clear geometric interpretations,
which can be precisely defined and measured using bbox coordinates of the objects.
In particular,
by extracting the bbox using an off-the-shelf object detector and tracker~\cite{ren2024grounded},
we compute the relative positions of the objects over time,
and derive a \ourmatric that quantifies the degree of compliance with the DSR instructions.

Our second contribution is to introduce a novel training strategy,
\textbf{\method},
to enhance the DSR capability of pre-trained T2V models.
A na{\"\i}ve approach would be to perform supervised fine-tuning (SFT) on real videos containing DSR scenarios.
However,
even if we collected such datasets,
the SFT approach does \emph{not} explicitly encourage the model to improve DSR alignment.
The model may simply memorize training videos,
without truly understanding the underlying SR.
Instead,
inspired by recent success in reinforcement learning from human feedback (RLHF)~\cite{ouyang2022training,shao2024deepseekmath,rafailov2023direct},
we propose to leverage \ourmatric to provide feedback signals for self-improvement of the T2V model.
In particular,
we adopt \textbf{D}irect \textbf{P}reference \textbf{O}ptimization (DPO)~\cite{rafailov2023direct} to fine-tune the model on the generated videos labeled and paired with the \ourmatric.
Our motivation for using DPO is two-fold:
(1) \ourmatric is a non-differentiable numeric signal,
which is not suitable for providing the direct gradient for SFT;
(2) online RL methods such as PPO~\cite{schulman2017proximal} or GRPO~\cite{shao2024deepseekmath} are computationally expensive,
due to the need for doing multi-step diffusion inference online.
By combining \ourmatric with DPO,
we effectively provide a scalable and efficient way to enhance the DSR capability of T2V models,
even without relying on real videos.

Finally,
to evaluate our method,
we built a new challenging benchmark \textbf{\ourdataset} of controlled DSR scenarios with diverse SR and motion patterns.
We also conducted extensive experiments on multiple state-of-the-art T2V models,
including
CogVideoX~\cite{yang2024cogvideox},
LTX-Video~\cite{hacohen2024ltx},
OpenSora~\cite{opensora2},
Wan2.1~\cite{wan2025} and
HunyuanVideo 1.5~\cite{hunyuanvideo2025}.
Our results demonstrate that our proposed \ourmatric is more reliable than VLM-based metrics in evaluating DSR correctness,
and our DPO-based training strategy effectively improves the DSR capability of T2V models,
surpassing baseline methods by a large margin.

Although our study focuses on DSR,
the problem formulation and solution are not limited to this specific task alone.
In fact,
our training strategy and evaluation metric offer greater value for physically-grounded video generation.
The geometry-based formulation underlying the \ourmatric
provides a general recipe for converting complex relational requirements into continuous, automatically computable signals,
useful for a wide range of applications.

In summary,
our key contributions are:
(1) \textbf{\ourmatric},
a geometry-based metric for reliable evaluation of DSR in generated videos,
which is more accurate and fine-grained than prior VLM-based approaches;
(2) \textbf{\method}, a DPO-based training strategy leveraging \ourmatric for aligning T2V models with DSR instructions,
which outperforms SFT and other baselines significantly;
and
(3) \textbf{\ourdataset}, a new benchmark dataset for controlled evaluation of DSR in T2V models,
along with extensive experiments demonstrating the effectiveness of our approach.

\section{Related Work}
\label{sec:related}

\begin{figure*}[t]
\centering
\includegraphics[width=\linewidth]{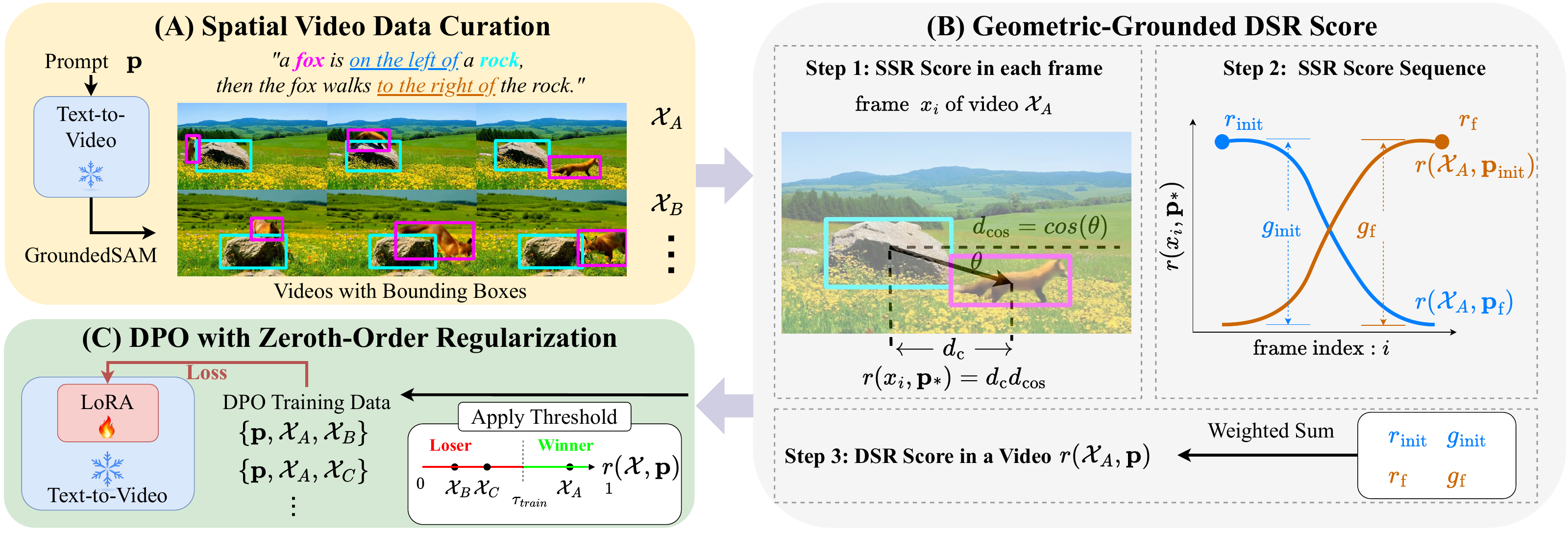}
\caption{\textbf{Overview of \method}.
\textbf{(A)} Given a text prompt,
the pre-trained T2V model generates several video samples.
For each sample, we first use GroundedSAM to obtain the bboxes of the animal and the object in each frame.
\textbf{(B)} Then, for each frame, we compute the Static Spatial Relationship (SSR) Score based on the bboxes.
From the SSR Score Sequence (all frames), we derive four metric components that are aggregated into the \ourmatric, which quantifies how well the video aligns with the dynamic spatial relationship (DSR).
\textbf{(C)} During DPO training, we identify winner/loser pairs based on the \ourmatric using a threshold.
We then train a LoRA to enhance the model's ability to accurately represent DSR in generated videos, using our proposed zero-order regularized DPO.
}
\label{fig:pipeline}
\end{figure*}


\paragraph{Spatial Reasoning in Vision.}
Early works,
like IQA~\cite{gordon2018iqa} and VQA~\cite{wu2017visual},
have explored spatial relationships in the context of visual question answering (VQA) tasks.
More recently,
SpatialVLM~\cite{chen2024spatialvlm} and SpatialReasoner~\cite{ma2025spatialreasoner} have focused on enhancing spatial reasoning capabilities in VLMs by incorporating 3D spatial annotations and explicit 3D representations, respectively.
To evaluate the spatial reasoning abilities of VLMs,
SpatialBench~\cite{xu2025spatialbench} has proposed a hierarchy of ability-oriented metrics.
In the realm of T2V generation,
the VBench series~\cite{huang2024vbench,zheng2025vbench} has introduced benchmarks for assessing DSR in generated videos.
However,
these evaluations primarily rely on vanilla VLMs,
which are not reliable enough~\cite{xu2025spatialbench}.
Instead,
we build a geometric-based evaluator by explicitly modeling  spatial relationships.

\vspace{-1em}
\paragraph{Spatial Alignment in T2I and T2V Generation.}
Like our approach,
spatially controlled generators aim to improve the spatial alignment between generated content and input prompts.
GLIGEN~\cite{li2023gligen} pioneered this direction by introducing a gated self-attention layer to ground entities within specified bboxes.
LayoutGPT~\cite{feng2023layoutgpt} employs a training-free method that translates prompts into webpage code-like formats to dictate object layouts.
BoxDiff~\cite{xie2023boxdiff} leverages text-to-latent cross-attention to enforce layout constraints during the denoising process.
VideoTetris~\cite{tian2024videotetris} decomposes prompts into frame-wise sub-prompts and region masks to enhance spatio-temporal consistency.
DyST-XL~\cite{he2025dyst} utilizes LLMs to interpret prompts and plan bboxes layouts, but its techniques are limited to specific architectures like MMDiT~\cite{esser2024scaling}.
Other works~\cite{qi2024layered,wu2024self,yang2024mastering,wang2025towards,lian2023llm} directly modify latent representations to achieve desired layouts.
However,
they often depends on auxiliary controls, such as bboxes, masks, or layouts, to guide the SR.
More importantly,
most of these focus on ``static'' T2I generation,
instead of ``dynamic'' T2V generation.

\paragraph{Preference Alignment in Generative models.}
Our method draws inspiration from DPO~\cite{rafailov2023direct} and its diffusion adaptation Diffusion-DPO~\cite{wallace2024diffusion}, which optimize generators using paired preference data.
These approaches bypass the need for online sampling,
such as in PPO~\cite{schulman2017proximal} and GRPO~\cite{shao2024deepseekmath}.
making them computationally efficient.
More recently,
this line of work has been extended to various diffusion-based generative models,
including
image generation~\cite{na2025boost,croitoru2025curriculum,karthik2025scalable},
video generation~\cite{liu2025videodpo,yang2025ipo,cai2025phygdpo},
and 3D asset generation~\cite{li2025dso}.
Our work is the first to apply DPO-based preference alignment to improve DSR reasoning in T2V generation.
\section{Method}
\label{sec:Method}

Given a pre-trained T2V generator
$p_\text{ref}$
that takes a text prompt
\textbf{p}
as input and generates a video
$\mathcal{X}_0 \sim p_\text{ref}(\mathcal{X}_0|\textbf{p})$,
our goal is to learn a new T2V model
$p_{\theta}$
that can generate
videos with subjects moving in a manner
that is more consistent with the dynamic spatial relationships described in the prompt
\textbf{p},
than the reference model
$p_\text{ref}$.

As illustrated in~\cref{fig:pipeline},
our work consists of three key components:
(1) we first collect a curated set of DSR prompts
and generate videos using the reference T2V model
$p_\text{ref}$;
(2) we then define and calculate \ourmatric,
a novel metric to determine whether a video
$\mathcal{X}_0$
is aligned with the DSR described in the prompt
\textbf{p};
(3) finally, we fine-tune the T2V model
$p_{\theta}$
using DPO~\cite{rafailov2023direct},
based on the preferences induced by \ourmatric.

\subsection{Problem Definition and Preliminaries}
\label{sec:preliminaries}

\paragraph{Problem Definition and Notations.}
To simplify the problem,
we focus on the DSR
between an
\emph{animal}
and a
\emph{static object}.
For example,
the sentence structure of a DSR prompt consists of a $\langle \text{\emph{scene}} \rangle$, an $\langle \text{\emph{animal}} \rangle$, a $\langle \text{\emph{static object}} \rangle$, the $\langle \text{\emph{initial static spatial relationship (SSR)}} \rangle$ and also the $\langle \text{\emph{final SSR}} \rangle$,
as shown in~\cref{fig:pipeline}.
We present ~\cref{tab:DSR_type} to give a comprehensive overview of the initial/final SSRs and the corresponding \emph{DSR type}.
We further introduce a set of abstract notations -- $\{\text{LEFT}, \text{RIGHT}, \text{TOP}\}$ -- as simplified representations for the ``on the xxx of" DSR keyword phrases used in the following of the paper for brevity.
In this paper, we explore DSRs formed by three types of SSRs: LEFT, RIGHT, and TOP, but our method is not specific to these and can be generalized to other types of DSRs.

\begin{table}[t]
    \centering
    \setlength{\tabcolsep}{6pt}
    \begin{tabular}{@{}llll@{}}
      \toprule
       & \textbf{DSR Type} & \textbf{Initial SSR} $\textbf{p}_\text{init}$ & \textbf{Final SSR} $\textbf{p}_\text{f}$ \\
       \midrule
       $\mathbb{A}$ & left-to-top   & on the left of  & to the top of   \\
       $\mathbb{B}$ & top-to-left   & on the top of   & to the left of  \\
       $\mathbb{C}$ & right-to-top  & on the right of & to the top of   \\
       $\mathbb{D}$ & left-to-right & on the left of  & to the right of \\
       $\mathbb{E}$ & top-to-right  & on the top of   & to the right of \\
       $\mathbb{F}$ & right-to-left & on the right of & to the left of  \\
       \bottomrule
    \end{tabular}
    \caption{\textbf{Overview} of DSR types and the corresponding keywords.
    }
    \label{tab:DSR_type}
\end{table}
In each video
$\mathcal{X}_0 = \{{\bm x}_i\}_{i=1}^{N}$,
where $N$ is the number of video frames,
we define the DSR as a transition from an initial static spatial relationship (SSR)
to a final SSR, as summarized in~\cref{tab:DSR_type}.
Specifically,
for each frame
${\bm x}_i$,
a corresponding SSR is defined as the spatial relationship between the
$\langle \text{\emph{animal}} \rangle$
and the
$\langle \text{\emph{static object}} \rangle$,
which is common in \emph{static} spatial T2I generators~\cite{li2023gligen,feng2023layoutgpt}.
Due to the movement of the animal,
the SSR will change over time.
Although such a change is described in the prompt,
the specific trajectory of the animal is \emph{not} strictly defined, allowing for substantial variability.

Therefore,
a key challenge in achieving DSR-aligned video generation is that it is difficult to define a precise optimization objective.
In contrast,
it is easy to determine whether a video
$\mathcal{X}_0$
is aligned with the DSR described in the prompt
\textbf{p}
by observing the change in SSR over time.
Hence,
we reformulate the problem as a \emph{preference optimization task},
where the preference is induced by the alignment level with the DSR,
which is quantified by our proposed \ourmatric.

\paragraph{Diffusion-DPO.}
As mentioned above,
 our goal is to optimize the T2V model
$p_{\theta}$
based on the preferences induced by the reward score.
The key idea is analogous to the T2I generator alignment in Diffusion-DPO~\cite{wallace2024diffusion}.
It optimizes the diffusion model using the preference data,
which is achieved by minimizing the following objective:
\begin{equation}
    \label{eq:dpo}
    \begin{aligned}
        \mathcal{L}_\text{DPO} = {} & - \mathbb{E}_{({\bm x}_0^w, {\bm x}_0^l)\sim D, t\sim\mathcal{U}(0, T), {\bm x}_t^w\sim q({\bm x}_t^w | {\bm x}_0^w), {\bm x}_t^l\sim q({\bm x}_t^l | {\bm x}_0^l)} \\
        & \log \operatorname{sigmoid} \bigg (- \beta T w (\lambda_t)) \Big( \\
        & \quad \Vert \bm{\epsilon}^w - \bm{\epsilon}_\theta(\bm{x}_t^w, t)\Vert ^2 _2 - \Vert \bm{\epsilon}^w - \bm{\epsilon}_\text{ref}(\bm{x}_t^w, t) \Vert^2_2 \\
        & \quad -\left( \Vert \bm{\epsilon}^l - \bm{\epsilon}_\theta(\bm{x}_t^l, t) \Vert ^2_2 - \Vert \bm{\epsilon}^l - \bm{\epsilon}_\text{ref}(\bm{x}_t^l, t) \Vert ^2_2 \right) \Big) \bigg) ,
    \end{aligned}
\end{equation}
where
${\bm x}_t^* = \alpha_t {\bm x}_0^* + \sigma_t {\bm \epsilon}^*$,
with
${\bm \epsilon}^* \sim \mathcal{N}(0, {\bm I})$,
and
$* \in \{w, l\}$
denotes the ``winner'' and ``loser'' samples, respectively.
Here,
$\bm{\epsilon}_\theta$
and
$\bm{\epsilon}_\text{ref}$
are the noise prediction networks of the fine-tuned and reference diffusion models, respectively.
$T$
is the total number of diffusion steps,
and
$\beta$
is a hyperparameter that controls the strength of the preference optimization.
$w(\lambda_t)$
is an important weight function.
Please refer to~\cite{wallace2024diffusion} for more details.

\subsection{Geometric-Grounded \ourmatric}
\label{sec:dsr_score}
The key to the success of DPO training is to define a reliable reward metric that can plausibly reflect the level of alignment with the DSR described in the prompt.
Existing spatial reasoning works~\cite{ma2025spatialreasoner,ma20253dsrbench,cai2025spatialbot} often rely on VLMs~\cite{hurst2024gpt,Qwen2.5-VL} to evaluate the correctness of spatial relationships.
However, as we will discuss in~\cref{sec:vlm_analysis}, such a VLM-based evaluation can be unreliable and inaccurate.
To address this issue,
we propose a novel geometrically based metric, termed \ourmatric,
to quantify the alignment level with the DSR.

\paragraph{Static Spatial Relationship (SSR) Score in each frame.}
Given a video
$\mathcal{X}_0 = \{{\bm x}_i\}_{i=1}^{N}$,
we first define the SSR-Score
$r({\bm x}_i, \textbf{p}_\text{*}) \in [-1, 1]$
to quantify the alignment level of each frame
${\bm x}_i$
with a specific SSR
$\textbf{p}_\text{*}$
expressed by the prompt
\textbf{p},
where 
$\textbf{p}_\text{*} \in \{\text{LEFT}, \text{RIGHT}, \text{TOP}\}$.

As illustrated in~\cref{fig:pipeline} (b) on the left,
the SSR-Score is computed based on the spatial relationship between the bboxes
$b_\text{a}$
and
$b_\text{o}$
 of the
animal
and the
static object,
which are obtained using GroundedSAM~\cite{ren2024grounded}.
Then,
the SSR-Score
is defined as the product of two quantities:
\begin{equation}
    \label{eq:ssr}
    r({\bm x}_i, \textbf{p}_\text{*}) = d_\text{c}(b_\text{a}, b_\text{o}) \: d_{\cos}(b_\text{a}, b_\text{o}),
\end{equation}
where
$d_\text{c}(b_\text{a}, b_\text{o}) = \operatorname{clip}(|\frac{b_\text{a}^{c_*}-b_\text{o}^{c_*}}{(b_\text{a}^{*} + b_\text{o}^{*})/2}|, 0, 1)$
is the normalized distance between the centers of the two bboxes
along the relevant axis
($*$ is the placeholder that takes ``$x$'' for LEFT/RIGHT and ``$y$'' for TOP,
$b_\text{*}^{c_*}$ denotes the center coordinate along the relevant axis,
and
$b_\text{*}^{*}$ denotes the width/height of the bbox along that axis);
and $d_{\cos}(b_\text{a}, b_\text{o})$
is the cosine distance between the object-to-animal center-based vector
$v(b_\text{a}, b_\text{o})$ and the axis.
The second term is necessary to ensure that the two entities are indeed positioned in the specified spatial relationship.
For example,
for the LEFT SSR,
the object-to-animal vector
should predominantly be aligned in the horizontal direction,
with a relatively small vertical component.

In a video generated with a prompt defining an initial SSR $\textbf{p}_\text{init}$ and a final SSR $\textbf{p}_\text{f}$, the actual SSR for each successive frame should change. To better understand this transition, we can use the SSR-Score to separately assess the alignment of each frame to both the initial and final prompted SSRs.
For example, suppose $\textbf{p}_\text{init}$=LEFT and $\textbf{p}_\text{f}$=TOP, then if our SSR-Score reports
$r({\bm x}_i, \text{LEFT})=0.7$, it
means that
${\bm x}_i$
satisfies LEFT at the level of 0.7,
while if
$r({\bm x}_i, \text{TOP})=0.2$, it
means that
${\bm x}_i$
satisfies TOP at the level of 0.2.
This
approach
helps to generalize the understanding of SSR to the whole video.
Specifically,
the alignment to SSR
$\textbf{p}_\text{*}$
of video
$\mathcal{X}_0$
can be described by the \emph{SSR-Score sequence}
$\{r({\bm x}_i, \textbf{p}_\text{*})\}_{i=1}^{N}$,
which is a list of per-frame SSR-Scores.

\paragraph{Dynamic Spatial Relationship (DSR) Score in one video.}
The \ourmatric
$r(\mathcal{X}_0, \textbf{p})$ is then defined based on the SSR-Scores sequence over all frames in
$\mathcal{X}_0$.
As illustrated in~\cref{fig:pipeline} (b) right,
an ideal video sample should exhibit a ``crossing'' pattern between the SSR-Score curves of the initial and final SSRs over time.
This is because,
as the animal moves according to the DSR,
the alignment level with the initial SSR
$\textbf{p}_\text{init}$
should decrease,
while the alignment level with the final SSR
$\textbf{p}_\text{f}$
should increase.
To capture this intuition,
the \ourmatric is defined
as follows:
\begin{equation}
    \label{eq:dsr}
    r(\mathcal{X}_0, \textbf{p}) = 0.125 (r_\text{init} + r_\text{f} + g_\text{init} + g_\text{f}) + 0.5 ,
\end{equation}
where
$0.125$
and $0.5$
are used for normalization to ensure that the \ourmatric lies in
$[0, 1]$,
where a higher score indicates better alignment with the DSR.
Here,
$r_\text{init} = \frac{1}{m}\sum_{k=1}^{m}r({\bm x}_k, \textbf{p}_\text{init})$
and
$r_\text{f} = \frac{1}{m}\sum_{k=N-m+1}^{N}r({\bm x}_k, \textbf{p}_\text{f})$
are the average SSR-Scores of the first and last
$m$ frames,
respectively,
which reflect the alignment quality with the initial and final SSRs at the two ends of the video. Additionally,
$g_\text{init} = r({\bm x}_1, \textbf{p}_\text{init}) - r({\bm x}_N, \textbf{p}_\text{init})$
and
$g_\text{f} = r({\bm x}_N, \textbf{p}_\text{f}) - r({\bm x}_1, \textbf{p}_\text{f})$
are the absolute differences between the SSR-Scores of the start and end frames,
which reflects the extent of transition.

\subsection{Spatial-Aligned Video Data Curation}
\label{sec:data_curation}
As shown in~\cref{fig:pipeline} (a),
we first generate samples using the reference T2V model
$p_\text{ref}$.
Before we reward these videos using our \ourmatric,
we first curate these samples to ensure that they are \textbf{VALID}.
Specifically,
we use a grounded object tracking model~\cite{ren2024grounded}
to track the bboxes of the animal and the static object in each generated video.
We then filter out the video samples that do not satisfy the following criteria:
(1) There is only \textbf{One} animal and \textbf{One} describing static object as designated in the prompt;
and (2) The two entities are successfully detected in at least 20 frames.
Then,
we compute the \ourmatric
$r(\mathcal{X}_0, \textbf{p})$
for each valid video
$\mathcal{X}_0$.
We use a threshold $\tau_\text{train}$
to divide the video samples under each prompt
\textbf{p},
where the samples with
$r(\mathcal{X}_0, \textbf{p}) \geq \tau_\text{train}$
are labeled as winners, otherwise losers.
Note that,
the existence of invalid samples is also an important reason why online RL methods such as PPO and GRPO are not applicable,
since their reward on an invalid sample can not provide meaningful signal for optimization.

\subsection{DPO with Zeroth-Order Regularization}
\label{sec:training_loss}

We can now fine-tune the T2V model
$p_{\theta}$
with the curated data
using the DPO loss defined in~\cref{eq:dpo}.
In particular,
we treat the positive samples as the ``winner''
samples
$\mathcal{X}_0^w$
and the negative samples as the ``loser''
samples
$\mathcal{X}_0^l$
for each prompt
\textbf{p}.
We follow the LoRA technique
to efficiently fine-tune the T2V model.
However, directly applying DPO training to our situation leads to degraded results, which is termed \emph{likelihood displacement}~\cite{pal2024smaug,pang2024iterative,razinunintentional}.
As shown in~\cref{fig:loss_curve} in Appendix~\ref{sec:loss_curve}, the DPO loss satisfies the margin by degrading the performance on both the winner and the loser, by forcing a much poorer performance on the losing side.

\begin{figure}[tb!]
\centering

\includegraphics[width=\linewidth]{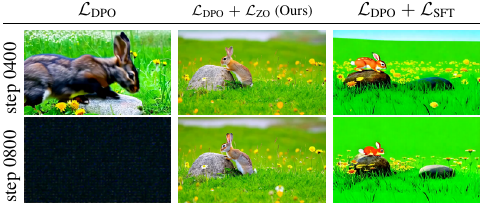}

\caption{\textbf{Ablation on loss}.
With only the DPO loss $\mathcal{L}_\text{DPO}$, the model degrades after 800 steps.
With SFT loss $\mathcal{L}_\text{SFT}$, the color saturation is strong. With our zeroth-order regularization $\mathcal{L}_\text{ZO}$, the training is stable.}
\label{fig:ablation_loss_visual}
\end{figure}



We attribute this issue to the nature of the DSR problem. Essentially, DSR are high-level concepts,
which are not directly reflected in the pixel values (in comparison, other properties such as aesthetics may be more closely linked to pixel values). Purely using DPO loss may lead the model to learn \emph{shortcuts} to satisfy the margin in an unintended way.
To address this issue,
we introduce an additional loss term to further regularize the fine-tuned model.
First,
we na\"ively combine the supervised fine-tuning (SFT) loss with the DPO loss:
\begin{equation}
    \label{eq:dpo_sft}
    \mathcal{L} = \mathcal{L}_\text{DPO} + \lambda_\text{SFT} \mathcal{L}_\text{SFT} ,
\end{equation}
where
$\mathcal{L}_\text{SFT}=
\Vert \bm{\epsilon}^w - \bm{\epsilon}_\theta(\mathcal{X}_t^w, t)\Vert ^2 _2 + \Vert \bm{\epsilon}^l - \bm{\epsilon}_\theta(\mathcal{X}_t^l, t)\Vert ^2 _2
$,
and $\lambda_\text{SFT}$ is a hyperparameter.
This simple combination helps to stabilize the training process and improve the DSR alignment,
but leads to sub-optimal results with over-saturation issues (as shown in~\cref{fig:ablation_loss_visual}).
To further mitigate this issue,
we introduce a \emph{Zeroth-Order} regularization term in addition to the \emph{First-Order} DPO loss,
\begin{equation}
    \label{eq:dpo_sft_zo}
    \mathcal{L} = \mathcal{L}_\text{DPO} + \lambda_\text{ZO} \mathcal{L}_\text{ZO} ,
\end{equation}
where
$\mathcal{L}_\text{ZO} =
\Vert \bm{\epsilon}_\text{ref}(\mathcal{X}_t^w, t) - \bm{\epsilon}_\theta(\mathcal{X}_t^w, t)\Vert ^2 _2 + \Vert \bm{\epsilon}_\text{ref}(\mathcal{X}_t^l, t) - \bm{\epsilon}_\theta(\mathcal{X}_t^l, t)\Vert ^2 _2
$,
and $\lambda_\text{ZO}$ is a hyperparameter.
This regularization considers the reference model as an anchor point,
which avoids the ``reward hacking'' issue where \emph{shortcuts} are learned to reduce preference loss,
but actually leads to the generation quality deviating excessively from those produced by the reference model.

\subsection{Reliability of VLM for Assessing DSRs}
\label{sec:vlm_analysis}


To better understand the advantages of our proposed \ourmatric,
we tested with Qwen3-VL-8B-Instruct~\cite{qwen3technicalreport} as the VLM-based metric~\cite{ma20253dsrbench} for DSR evaluation, in accordance with popular practice.
Following VBench-2.0~\cite{zheng2025vbench},
we designed two questions for the initial SSR and the final SSR, respectively:

(1) \textit{Is the $\langle$animal$\rangle$ initially $\langle$initial SSR$\rangle$ the $\langle$object$\rangle$? Answer yes or no.}

(2) \textit{Is the $\langle$animal$\rangle$ finally $\langle$final SSR$\rangle$ the $\langle$object$\rangle$? Answer yes or no.}

We collected the YES/NO answers given by the VLM and correlate them with the SSR-Scores of the initial SSR and the final SSR on a large set of videos.

\begin{figure}[tb!]
\centering
\includegraphics[width=\linewidth]{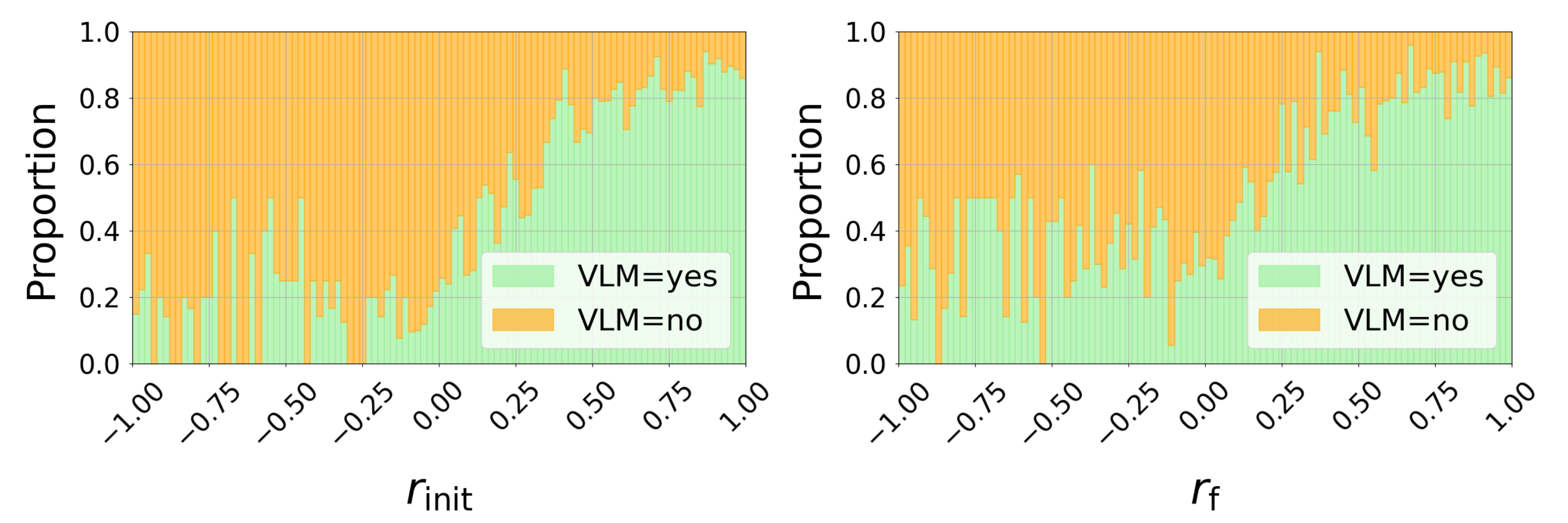}
\vspace{-2em}
\caption{
\textbf{VLM \emph{vs} \ourmatric.}
Each plot contains bins at each SSR Score value.
Each bin shows the portion where the VLM gives the \textcolor{green}{YES} answer or \textcolor{orange}{NO} answer at the particular SSR Score location.
The VLM gives a significant portion of \textcolor{green}{YES} on the low SSR Score interval, especially for the evaluation on the final SSR.}
\label{fig:VLM}
\end{figure}
\Cref{fig:VLM} shows the correlation plots of VLM answers \emph{vs} {\ourmatric}s.
As formulated in~\cref{sec:dsr_score}, $r_\text{init}$ / $r_\text{f}$ reflects the level of alignment to the initial/final SSR of the video.
However, the plots show that the VLM gives a non-negligible portion of YES answers on videos with low scores.
This surprisingly shows that the VLM tends to provide positive responses regardless of the actual spatial relationship in the video.
Nonetheless, we comparatively evaluated the use of VLM answers as the reward for training, and the details are discussed in~\cref{sec:ablations}.

\section{Experiments}
\label{sec:experiment}

\begin{figure*}[t]
\centering

\includegraphics[width=\linewidth]{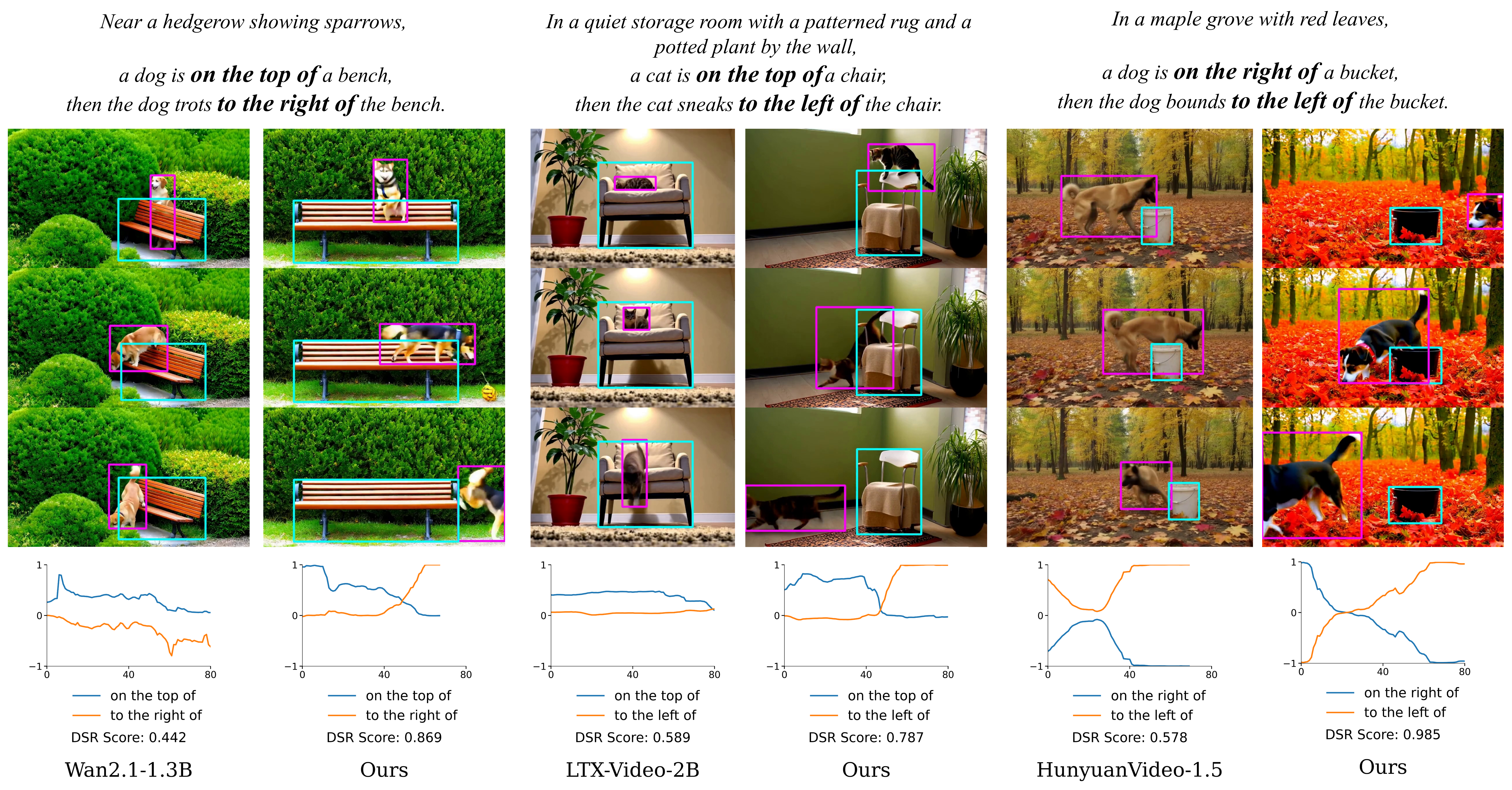}

\caption{\textbf{Qualitative comparisons with the state-of-the-art T2V models on \ourdataset.}
Our fine-tuned model (right columns) generates videos that correctly reflect the desired change in spatial relationships,
while the baseline models (left columns) fail to do so.}
\label{fig:qualitative_results_set0_v2}
\end{figure*}

\subsection{Settings}
\label{sec:experiment_settings}

\paragraph{Dataset.}
We conducted experiments on our curated dataset named \ourdataset,
designed for training and evaluating T2V models on DSR prompts.
For trianing,
we collected 500 DSR prompts using GPT-4o,
each accompanied by 10 video samples generated with random seeds.
For testing,
we curated 120 DSR prompts,
each evaluated with 5 fixed seeds to ensure reproducibility.
Note that,
(1) only \textbf{VALID} samples are involved in the training;
and (2) the exact \ourmatric value is not involved for DPO training.
More details about the dataset are provided in Appendix~\ref{sec:dataset}.

\paragraph{Metrics.}
For quantitative results,
we report the \textbf{Correctness@0.7} metric,
ID Consistency,
CLIP-IQA~\cite{wang2023exploring},
and Imaging Quality.
Correctness@0.7 indicates the percentage of generated videos achieving an \ourmatric value
greater than or equal to 0.7,
reflecting the model's ability to align with DSR prompts.
ID Consistency measures the consistency of the animal appearance along the video frames,
by extracting DINOv2 features~\cite{oquab2023dinov2} on the bbox area containing the animal.
CLIP-IQA and Imaging Quality collectively assess the visual quality of the generated samples.
Limited by space,
more details about the metrics are provided in~\cref{sec:supp_experiment_settings}.

\begin{table}[t]
\centering
\resizebox{\linewidth}{!}{
\begin{tabular}{@{}l cccc @{}}
    \toprule
    \textbf{Method} & \textbf{Correct@0.7} $\uparrow$ & \textbf{ID} $\uparrow$ & \textbf{CLIP-IQA} $\uparrow$ & \textbf{IQ} $\uparrow$ \\
    \midrule
    CogVideoX1.5-5B        & 0.053             & 0.6404 & 0.8874 & 0.5894 \\
    OpenSora v1.2          & 0.018             & 0.7151 & 0.8712 & 0.6616 \\
    LTX-Video-2B           & 0.058             & \textbf{0.8028} & \underline{0.9061} & 0.6786 \\
    HunyuanVideo v1.5-8B   & \underline{0.490} & \underline{0.7565} & \textbf{0.9299} & 0.6662 \\
    Wan2.1-1.3B            & 0.125             & 0.7046 & 0.8481 & \textbf{0.7000} \\
    Wan2.1-1.3B + LoRA (Ours) & \textbf{0.585} & 0.6934 & 0.8243 & \underline{0.6909} \\
    \bottomrule
\end{tabular}
}
\caption{\textbf{Quantitative Results on \ourdataset}.
``Correct@0.7'' denotes Correctness at threshold 0.7.
``ID'' denotes ID Consistency.
``CLIP-IQA'' denotes CLIP-based Image Quality Assessment.
``IQ'' denotes Imaging Quality.
The best and second-best results are \textbf{bolded} and \underline{underlined}, respectively.
Our method achieves the best correctness while maintaining decent image quality.
}
\label{tab:Quantitative Results v2}
\end{table}



\paragraph{Implementation Details.}
\method is mainly trained with the Wan2.1-1.3B~\cite{wan2025} T2V model.
However, our method is architecture-agnostic
and can be applied to a wide range of I2V models.
We fine-tune two prompt-related components using LoRA~\cite{hu2022lora}:
(1) the projection module mapping text embeddings into the model’s latent space, and
(2) the key and value projection matrices in every cross-attention layer.
The video consists of 81 frames at 480$\times$832 resolution.
We use AdamW~\cite{loshchilov2017decoupled} to fine-tune the model with LoRA at rank 16.
The learing rate is 1e-4.
The batch size is 48 and the training is conducted for 2400 steps on 4 RTX4090's.
We set $\beta=1$ in the DPO loss and $\lambda_{ZO}=0.25$.
More details are provided in~\cref{sec:supp_experiment_settings}.

\subsection{Main Results}
\label{sec:main_results}

\paragraph{Quantitative Results.}
We compare \method to the state-of-the-art T2V models in~\cref{tab:Quantitative Results v2},
including
Open-Sora 2.0~\cite{opensora2},
CogVideoX1.5-5B~\cite{yang2024cogvideox},
Wan2.1-1.3B~\cite{wan2025},
LTX-Video-2B~\cite{hacohen2024ltx},
and HunyuanVideo1.5-8B~\cite{hunyuanvideo2025}.
\method significantly outperforms these baselines in Correctness@0.7, demonstrating its effectiveness in generating videos that are better aligned with DSR prompts.
The detailed \emph{correctness curve} is shown in~\cref{fig:curve_main},
showing how \method consistently improves correctness across various thresholds.
For the other metrics,
\method maintains comparable ID Consistency, visual quality and Image Quality to the baseline Wan2.1-1.3B,
demonstrating that the fine-tuning does not compromise these aspects.
Note that,
HunyuanVideo1.5-8B achieves the second best ID Consistency and the best CLIP-IQA score,
likely attributed to its larger model size.

\paragraph{Qualitative Results.}
\method outputs are visually compared with the latest T2V models in~\cref{fig:qualitative_results_set0_v2}.
\method generates videos that accurately reflect the spatial dynamics described in the DSR prompts,
while maintaining high visual fidelity.
In contrast,
the baseline 
and other SOTA models
often fail to capture the correct spatial relationships,
resulting in less coherent videos.




\begin{table*}[tb!]
    \subfloat[{
        \footnotesize \textbf{Training loss.} SFT with top-2 samples brings limited improvement.
        Removing $\tau$ degrades the performance significantly.}
        \label{tab:ablation_method}
    ]{
        \begin{minipage}{0.48\linewidth}{
            \begin{center}
                \resizebox{\linewidth}{!}{
                        \begin{tabular}{@{} l cccc @{}}
                            \toprule
                            \textbf{Method} & \textbf{Correct@0.7} $\uparrow$ & \textbf{ID} $\uparrow$ & \textbf{CLIP-IQA} $\uparrow$ & \textbf{IQ} $\uparrow$ \\
                            \midrule
                            SFT, top-2                      & 0.187 & \textbf{0.7549} & 0.6330 & 0.7113 \\
                            \midrule
                            DPO, no $\tau_{train}$          & 0.113 & 0.7102 & \textbf{0.8667} & \textbf{0.7298} \\
                            $\mathcal{L}_\text{DPO}+\mathcal{L}_\text{SFT}$, $\tau_{train}=0.7$ & \textbf{0.333} & 0.7317 & 0.6549 & 0.7107 \\
                            $\mathcal{L}_\text{DPO}+\mathcal{L}_\text{ZO}$ (Ours)  & 0.307 & 0.6754 & 0.8594 & 0.7141 \\
                            \bottomrule
                        \end{tabular}
                    }
            \end{center}
        }
        \end{minipage}
    }  
    \hspace{2em}
    \subfloat[{
        \footnotesize \textbf{Reward system.} 
        Our \ourmatric outperforms Qwen3-VL-8B and VBench-2.0 significantly on spatial correctness.
        }
        \label{tab:ablation_judge}
    ]{
        \begin{minipage}{0.47\linewidth}{
            \begin{center}
                \resizebox{\linewidth}{!}{
                        \begin{tabular}{@{} l cccc @{}}
                            \toprule
                            \textbf{Method} & \textbf{Correct@0.7} $\uparrow$ & \textbf{ID} $\uparrow$ & \textbf{CLIP-IQA} $\uparrow$ & \textbf{IQ} $\uparrow$ \\
                            \midrule
                            Qwen3-VL-8B-Instruct      & 0.147 & 0.7269 & 0.8746 & 0.7268 \\
                            VBench-2.0                & 0.147 & \textbf{0.7435} & 0.8738 & \textbf{0.7301} \\
                            \midrule
                            Endpoint SSR          & 0.180 & 0.7243 & \textbf{0.8750} & 0.7240 \\
                            \ourmatric (Ours)          & \textbf{0.307} & 0.6754 & 0.8594 & 0.7141 \\
                            \bottomrule
                        \end{tabular}
                    }
            \end{center}
        }
        \end{minipage}
    }
    \\[1em]
    \subfloat[{
        \footnotesize \textbf{Threshold $\tau_{train}$} for splitting winner/loser data. Large $\tau_{train}$ leads to better spatial correctness,
        but worse video quality.
        }
        \label{tab:ablation_trainTH}
    ]{
        \begin{minipage}{0.47\linewidth}{
            \begin{center}
                \renewcommand{\arraystretch}{1.2}
                \resizebox{\linewidth}{!}{
                        \begin{tabular}{@{} l cccc @{}}
                            \toprule
                            \textbf{Method} & \textbf{Correct@0.7} $\uparrow$ & \textbf{ID} $\uparrow$ & \textbf{CLIP-IQA} $\uparrow$ & \textbf{IQ} $\uparrow$ \\
                            \midrule
                            Wan2.1-1.3B (Baseline)            & 0.180          & 0.6935 & \textbf{0.8723} & 0.7293 \\
                            \midrule
                            0.6         & 0.200 & \textbf{0.7152} & 0.8635 & \textbf{0.7335} \\
                            0.7 (Ours)  & 0.307 & 0.6754 & 0.8594 & 0.7141 \\
                            0.8         & \textbf{0.367} & 0.6466 & 0.8401 & 0.7153 \\
                            \bottomrule
                        \end{tabular}
                    }
            \end{center}
        }
        \end{minipage}
    } 
    \hspace{2em}
    \subfloat[{
        \footnotesize \textbf{Different form of prompt.} 
        Our method consistently outperforms the baseline under different prompt forms.
        }
        \label{tab:ablation_prompt}
    ]{
        \begin{minipage}{0.47\linewidth}{
            \begin{center}
                \resizebox{\linewidth}{!}{
                        \begin{tabular}{@{}l l cccc @{}}
                            \toprule
                            \textbf{Augmentation} & \textbf{Method} & \textbf{Correct@0.7} $\uparrow$ & \textbf{ID} $\uparrow$ & \textbf{CLIP-IQA} $\uparrow$ & \textbf{IQ} $\uparrow$ \\
                            \midrule
                            \multirow{2}{*}{ChatGPT} & baseline      & 0.107 & 0.6621 & 0.8626 & 0.7176 \\
                            & Ours               & 0.407 & 0.6816 & 0.8461 & 0.7039 \\
                            \midrule
                            \multirow{2}{*}{Qwen2.5} & baseline        & 0.180 & 0.6994 & 0.8511 & 0.7033 \\
                            & Ours          & 0.433 & 0.7231 & 0.8402 & 0.6884 \\
                            \midrule
                            \multirow{2}{*}{\textit{from ... to ...}} & baseline  & 0.107 & 0.5967 & 0.8602 & 0.7125 \\
                            & Ours               & 0.367 & 0.6345 & 0.8423 & 0.6954 \\
                            \bottomrule
                        \end{tabular}
                    }
            \end{center}
        }
        \end{minipage}
    }
    \captionof{table}{\textbf{\method Ablations.}
    All models are fine-tuned for 800 steps unless specified otherwise.
    Results are evaluated on a subset of 30 test prompts.
    The best results are \textbf{bolded}.}
    \label{tab:ablation}
\end{table*}

\subsection{Ablations}
\label{sec:ablations}

We conducted thorough ablations on various components of our method, as summarized in~\cref{tab:ablation}.
Due to resource constraints,
all ablations were performed with 800 training steps
and on a subset of 500 training prompts and 30 testing prompts,
unless specified otherwise.

\paragraph{Training loss.}
The DPO training serves as a cornerstone to the success of \method,
as it is hard to directly supervise spatial relationship alignment.
To validate this,
we compare our default DPO training with two alternatives in \cref{tab:ablation_method}:
(1) supervised fine-tuning (SFT) using the top-2-\ourmatric samples,
and (2) DPO without a global threshold $\tau_\text{train}$,
where the winner/loser pairs are randomly sampled based on their relative \ourmatric values.
The results show that our DPO with a global threshold outperforms both alternatives,
highlighting the effectiveness of our training strategy.
Note that
SFT-based methods lead to significant degradation in visual quality,
as indicated by the low CLIP-IQA ``natural'' score,
even if supervised by high-\ourmatric samples.
We also compared the alternative SFT regularization loss to our zeroth-order regularization loss $\mathcal{L}_\text{ZO}$.
Note that
with more training steps (2,400 steps),
our DPO with $\mathcal{L}_\text{ZO}$ outperforms the SFT-based method in all metrics (\cref{fig:ablation_loss}).
More importantly,
the SFT-based supervision or regularization suffers from color saturation issues,
resulting in unrealistic video appearances (\cref{fig:ablation_loss_visual}).

\paragraph{Reward System.}
Our geometry-based \ourmatric is significant in providing a reliable reward system for DSR alignment.
We investigated by comparing with two VLM-based reward systems in \cref{tab:ablation_judge}:
(1) Qwen3-VL-8B-Instruct~\cite{qwen3technicalreport}, and
(2) VBench-2.0~\cite{zheng2025vbench}.
When trained with these VLM-based rewards,
the spatial correctness is not only significantly lower than using our \ourmatric,
but is even worse than the baseline without fine-tuning (\cref{tab:ablation_trainTH}),
indicating that VLM-based spatial judgment is unreliable for use as a reward signal.

\paragraph{\ourmatric Component.}
We further ablate the components of \ourmatric in \cref{tab:ablation_judge}.
Specifically, we remove the gap values $g_\text{init}, g_\text{f}$ from \ourmatric
and only use the endpoint values $r_\text{init}, r_\text{f}$ as reward scores.
The results show that removing the gap values leads to degraded performance,
indicating that including information on the transition process is important for overall performance enhancement.

\paragraph{Ablations on threshold $\tau_\text{train}$.}
The default threshold of $\tau_\text{train}=0.7$
is used for splitting winner/loser training samples.
We ablate the choice of $\tau_\text{train}$ in \cref{tab:ablation_trainTH} and \cref{fig:ablation_trainTH}.
With $\tau_\text{train}=0.6$,
the model is unable to significantly improve spatial correctness.
Conversely,
with $\tau_\text{train}=0.8$,
the model gives a higher performance on spatial correctness,
showing that a higher $\tau_\text{train}$ enhances the signal-to-noise ratio for winner/loser pairs.
Evidently our \ourmatric
is effective in assessing spatial alignment
and can serve as a reliable reward signal.
However,
a higher $\tau_\text{train}$ leads to a lower CLIP-IQA ``natural'' score,
indicating degraded naturalness in visual appearance.
This can be explained by the decreasing percentage of available training data,
since a higher $\tau_\text{train}$ filters out more winner training samples.
Thus,
there is a trade-off in selecting $\tau_\text{train}$.

\paragraph{Ablations on different prompt structure.}
Another important question is
whether the effectiveness of our method is sensitive to the prompt structure.
To test the prompt generalizability of our method,
we test our fine-tuned model on alternative structures of DSR prompts (\cref{tab:ablation_prompt}).
We tested three ablations:
(1) augmenting the prompt with ChatGPT~\cite{hurst2024gpt},
(2) likewise with Qwen2.5-7B-Instruct~\cite{Qwen2.5-VL},
and (3) modifying the prompt into a \textit{``from ... to ...''} structure,
such as \textit{``a fox is on the left of a chair, then the fox sprints to the right of the chair''} being rephrased as \textit{``a fox sprints from the left of a chair to the right of the chair''}.
For fair comparison,
we used the same model fine-tuned on \textbf{our structured prompt}
for testing on all three alternative prompt structures.
Across all these alternative prompts,
the fine-tuned model performs better than the baseline,
suggesting that the fine-tuned model has learned deeper  semantic knowledge about spatial relationships,
beyond simply overfitting to prompt structure.

\section{Conclusion}
\label{sec:conclusions}

We present \method,
a self-improvement framework for enhancing T2V generators in modeling dynamic spatial relationships (DSR).
The key to our success is that we construct a zero-order DPO reward function to fine-tune the T2V models towards better alignment with DSR. 
Unlike prior works that rely on VLMs for evaluating spatial relationships,
we propose a geometry-based \ourmatric metric to provide more reliable and interpretable measurement of DSR alignment.
The resulting model demonstrates a significant improvement in generating videos that accurately reflect the dynamic spatial relationships specified in the text prompts,
where preserve the identity and image quality.
The simplicity and efficiency of our approach also make it a potentially promising solution for enhancing T2V models in other high-level physical attributes beyond DSR.


\paragraph{Limitations and Discussions.}
Despite the promising results,
\method has some limitations.
First,
the calculation of the geometry-based \ourmatric is highly dependent on the robustness of GroundedSAM~\cite{ren2024grounded}.
However,
it does not always perform perfectly,
especially for generated videos with complex scenes and motion blur.
These defects can lead to incorrect \ourmatric values or even invalid samples.
A more robust detection and tracking tool should be adopted for a more reliable \ourmatric system.
Second, 
our work only explores the setting of one animal and one object with DSRs on LEFT, RIGHT, and TOP.
More complicated settings, such as a scene graph,
should be explored to prove the generalizability of our design of the reward on dynamic spatial relationships.

\paragraph{Acknowledgments.}
This research is supported by the Ministry of Education, Singapore, under its Academic Research Fund Tier 1 RG107/24. Chuanxia Zheng is supported by NTU SUG-NAP and the National Research Foundation, Singapore, under its NRF Fellowship Award
NRF-NRFF17-2025-0009. 


\bibliography{example_paper,chuanxia_general,chuanxia_specific}
\bibliographystyle{icml2026}


\newpage
\appendix
\renewcommand{\thesection}{A\arabic{section}}
\renewcommand{\thefigure}{A\arabic{figure}}
\renewcommand{\thetable}{A\arabic{table}}
\clearpage

\twocolumn[{%
  \begin{center}
    \vspace*{1em}
    {\LARGE\bfseries Supplemental Material}
    \vspace{2em}
  \end{center}
}]

\section{Additional Experimental Settings}
\label{sec:supp_experiment_settings}
\subsection{Metrics}
\textbf{Correctness}
We use the \ourmatric defined in~\cref{sec:dsr_score} to measure how well the generated video aligns with the dynamic spatial relationship (DSR) described in the prompt.
In particular,
we count the \emph{percentage} of samples in the test set having \ourmatric $>=\tau_\text{test}$ as the correctness (denoted as Correctness@$\tau_\text{test}$) of DSR.
Intuitively, it indicates the rate where the T2V model can produce DSR samples up to a certain level of quality, reflecting the model's general ability to align with DSR prompts.
At $\tau_\text{test}=0$, correctness is validness, reflecting the percentage of valid samples among the total number of test samples.
We particularly focus on Correctness@0.7, effectively having $\tau_\text{test}=\tau_\text{train}$.
In addition to the single Correctness@0.7 value, we also check the \emph{correctness curve}, which is the correctness at varied $\tau_\text{test} \in [0,1]$.

\textbf{ID Consistency}
We aim to measure the consistency of the animal appearance along the video frames.
VBench~\cite{huang2024vbench} measures subject consistency by checking the DINOv2~\cite{oquab2023dinov2} feature similarity of the frames with respect to the first frame and the previous frame.
However, the calculation is based on the whole frame, where the background is involved.
In order to have a more precise assessment of the consistency of the \emph{animal}, we only extract DINOv2 features within the bounding box containing the animal for the similarity measurement.

\textbf{CLIP-IQA and Imaging Quality}
We use CLIP-IQA~\cite{wang2023exploring} and Imaging Quality (from VBench~\cite{huang2024vbench}) to collectively assess the visual quality of the generated samples.
For CLIP-IQA, we compute at the ``natural'' mode on the video frames and take average.
A higher CLIP-IQA ``natural'' score means that the video frame is more likely classified as \emph{``Natural photo''} than \emph{``Synthetic photo''}.
For Imaging Quality, we follow VBench~\cite{huang2024vbench} to extract the DINOv2~\cite{oquab2023dinov2} feature for the whole frame, and compute the sharpness and colorfulness.
A higher Imaging Quality score means that the video frame is sharper and more colorful.
For both CLIP-IQA and Imaging Quality, we take the average score on all frames as the final score for the video.

\subsection{Examples of our curated \ourdataset}
\label{sec:dataset}
\begin{itemize}[itemsep=1pt, parsep=0pt]
\footnotesize
\item On a grassy field with wildflowers, a rabbit is \onleft a stone, then the rabbit jumps \toright the stone.
\item In a quiet forest clearing, a squirrel is \onleft a lamp, then the squirrel scampers \toright the lamp.
\item By a calm riverbank with reeds, a cat is \onleft a hydrant, then the cat paces \toright the hydrant.
\item At the edge of a sunny meadow, a dog is \onleft a bucket, then the dog runs \toright the bucket.
\item On a rocky hillside with moss, a fox is \onleft a chair, then the fox sprints \toright the chair.
\item At a seaside dock with gulls, a turtle is \onleft a bench, then the turtle ambles \totop the bench.
\item Near a market square with stalls, a rabbit is \onleft a crate, then the rabbit hops \totop the crate.
\item On a plaza with stone benches, a squirrel is \onleft a stone, then the squirrel darts \totop the stone.
\item Inside a hallway with framed photos, a cat is \onleft a lamp, then the cat sneaks \totop the lamp.
\item By a village well with buckets, a dog is \onleft a chair, then the dog trots \totop the chair.
\item On a breezy hilltop at dusk, a cat is \onright a hydrant, then the cat prowls \toleft the hydrant.
\item In a maple grove with red leaves, a dog is \onright a bucket, then the dog bounds \toleft the bucket.
\item By a pebble beach with driftwood, a fox is \onright a chair, then the fox dashes \toleft the chair.
\item At a fountain plaza with tiles, a bird is \onright a bench, then the bird flutters \toleft the bench.
\item Inside a sunroom with potted plants, a duck is \onright a crate, then the duck shuffles \toleft the crate.
\item At a gazebo with benches, a squirrel is \onright a stone, then the squirrel scampers \totop the stone.
\item On a slope with stepping stones, a cat is \onright a lamp, then the cat paces \totop the lamp.
\item By a trailhead signpost, a dog is \onright a chair, then the dog runs \totop the chair.
\item In a colonnade with pillars, a fox is \onright a bench, then the fox sprints \totop the bench.
\item At a riverside promenade with lamps, a bird is \onright a crate, then the bird pecks \totop the crate.
\item At a pergola featuring climbing vines, a monkey is \ontop a log, then the monkey leaps \toleft the log.
\item At a pergola showing climbing vines, a turtle is \ontop a bench, then the turtle shuffles \toleft the bench.
\item Inside a gallery featuring white walls, a rabbit is \ontop a stone, then the rabbit jumps \toleft the stone.
\item Inside a gallery showing white walls, a squirrel is \ontop a desk, then the squirrel scampers \toleft the desk.
\item On a courtyard deck featuring lanterns, a cat is \ontop a table, then the cat paces \toleft the table.
\item On a courtyard deck showing lanterns, a squirrel is \ontop a stone, then the squirrel darts \toright the stone.
\item Near a hedgerow featuring sparrows, a cat is \ontop a desk, then the cat sneaks \toright the desk.
\item Near a hedgerow showing sparrows, a dog is \ontop a bench, then the dog trots \toright the bench.
\item By a canal featuring brick edges, a fox is \ontop a chair, then the fox trots \toright the chair.
\item By a canal showing brick edges, a bird is \ontop a crate, then the bird hops \toright the crate.

\end{itemize}

\begin{figure}[tb!]
\centering

\begin{subfigure}[t]{0.48\linewidth}
    \centering
    \includegraphics[width=\linewidth]{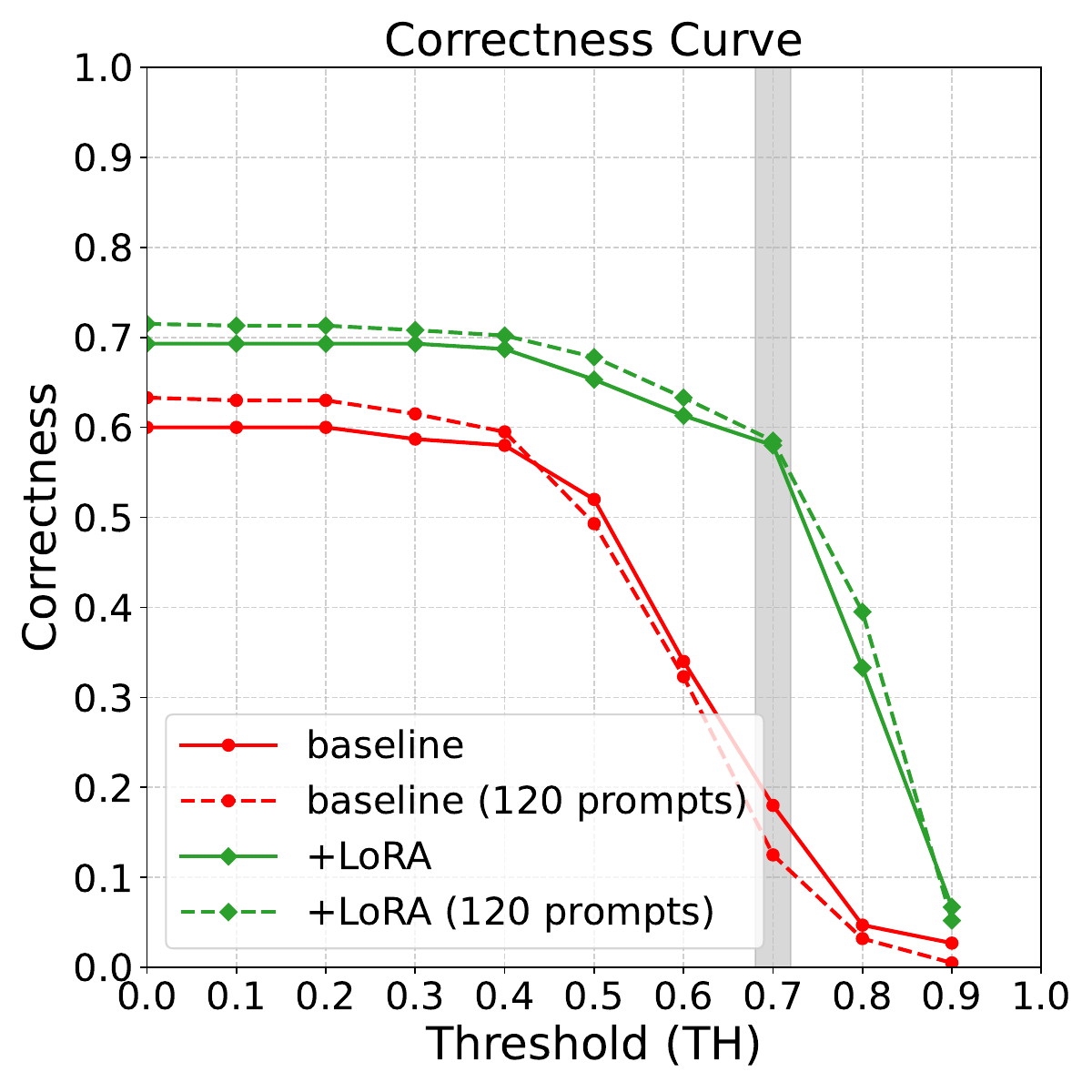}
    \caption{Main result on Wan2.1--1.3B}
    \label{fig:curve_main}
\end{subfigure}
\begin{subfigure}[t]{0.48\linewidth}
    \centering
    \includegraphics[width=\linewidth]{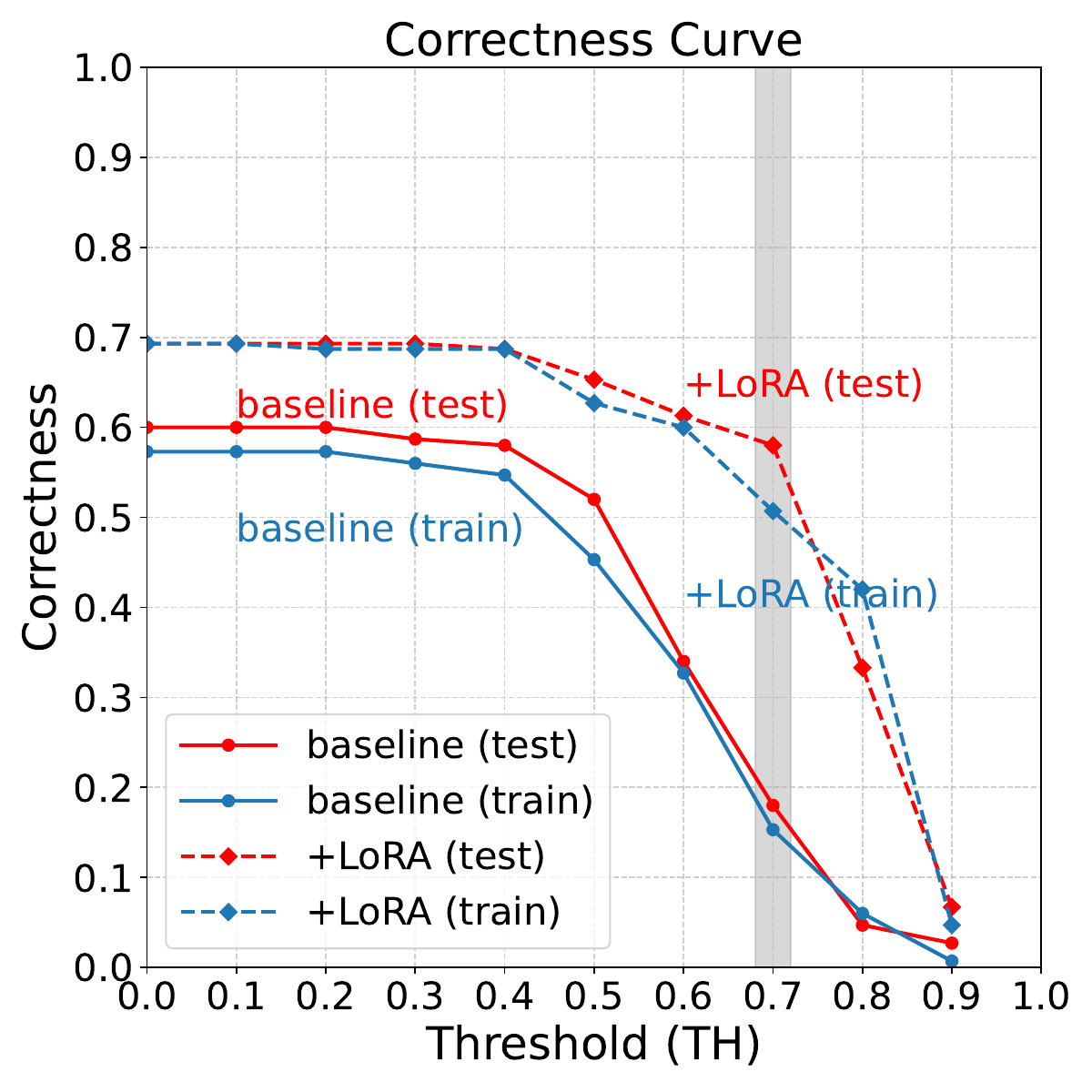}
    \caption{Test set vs. Train set (subset)}
    \label{fig:curve_overfitting}
\end{subfigure}

\caption{\textbf{Correctness curve}.
The percentage of samples having \ourmatric$>=\tau_\text{test}$ under different $\tau_\text{test}$.
The emphasized focus (grey area) is on $\tau_\text{test}=0.7$.
(a) The training significantly enhances the model's ability in aligning with DSR prompts. 
(b) We compare the result on the test set to a subset of training set (also 30 prompts), and indicate that the fine-tuned model is not overfit to the training set. }
\label{fig:correctness_curve}
\end{figure}
\section{Additional Analysis}
\label{sec:loss_curve}

\paragraph{Ablations on the training set.}
DPO is prone to overfitting due its offline nature of training on collected data, compared to PPO/GRPO which are trained on online data.
To investigate whether our fine-tuned model has overfit to the training set,
we selected a subset of 30 prompts from the training set for testing.
As shown in~\cref{fig:curve_overfitting},
the fine-tuned model demonstrates a similar performance enhancement on this training subset as that on the test set.
This indicates that the overfitting is not significantly observed and our method appears to have acquired semantic-level understanding of spatial relationships.

\paragraph{Analysis of Internal Changes in Text-to-Latent Attention.}
Apart from monitoring the performance enhancement,
it is also essential to analyze the internal changes caused by the fine-tuning.
Here we investigate the angle of \emph{text-to-latent} correlation.
Since the model has a classical \emph{cross-attention} design whereby  the text tokens (tokenized from the prompt) attend to the video latents,
we consider the patterns in the \emph{cross-attention activation map} (CAMAP) to be reflective of the text-to-latent correlation. 

In particular,
we focus on the \emph{similarity} between the CAMAP values (flattened as a vector) of the tokens for initial/final SSR with respect to the $\langle \text{\emph{animal}} \rangle$ and the $\langle \text{\emph{static object}} \rangle$ tokens.
This approach is taken because SSR is not a tangible concept like $\langle \text{\emph{animal}} \rangle$ or $\langle \text{\emph{static object}} \rangle$ that is directly reflected in the geometrical patterns within CAMAP, but rather has a modifier effect.
The similarity in CAMAP reflects the strength of binding between the SSR tokens and $\langle \text{\emph{animal}} \rangle$/$\langle \text{\emph{static object}} \rangle$.

As shown in~\cref{tab:camap_analysis},
we split the prompt into 5 groups and inspect the similarity of CAMAP values between the initial/final SSR group and the 5 groups.
Comparing the baseline and the fine-tuned model, significantly increased correlation occurs for $\langle \text{\emph{initial SSR}} \rangle\text{\emph{-to-}}\langle \text{\emph{animal}} \rangle$ and $\langle \text{\emph{final SSR}} \rangle\text{\emph{-to-}}\langle \text{\emph{animal}} \rangle$.
This provides some evidence that our method takes effect by making the tokens of initial/final SSR more tightly bound to the tokens of $\langle \text{\emph{animal}} \rangle$.

\begin{table}[tb!]
\centering
\resizebox{\linewidth}{!}{
\begin{tabular}{@{}ll cccccc @{}}
\toprule
& & Animal & Object & Initial SSR & Final SSR & Others \\
\midrule
\multirow{2}{*}{Baseline} & Initial SSR & 0.395 & 0.280 & 1.000 & 0.837 & 0.878 \\
& Final SSR   & 0.435 & 0.278 & 0.837 & 1.000 & 0.892 \\
\midrule
\multirow{2}{*}{+ LoRA} & Initial SSR & 0.456 & 0.277 & 1.000 & 0.866 & 0.894 \\
 & Final SSR   & 0.489 & 0.274 & 0.866 & 1.000 & 0.903 \\
\bottomrule
\end{tabular}
}
\caption{
\textbf{CAMAP Analysis}.
We compute the average CAMAP correlation between different types of tokens:
Animal-related tokens, Object-related tokens, Initial SSR-related tokens, Final SSR-related tokens, and Other tokens.
The fine-tuned model shows a stronger correlation between the $\langle \text{\emph{initial SSR}} \rangle\text{\emph{-to-}}\langle \text{\emph{animal}} \rangle$ and $\langle \text{\emph{final SSR}} \rangle\text{\emph{-to-}}\langle \text{\emph{animal}} \rangle$ pairs, indicating better capturing of the spatial relationships specified in the DSR prompts.
}
\label{tab:camap_analysis}
\end{table}

\begin{figure}[tb!]
\centering

\includegraphics[width=\linewidth]{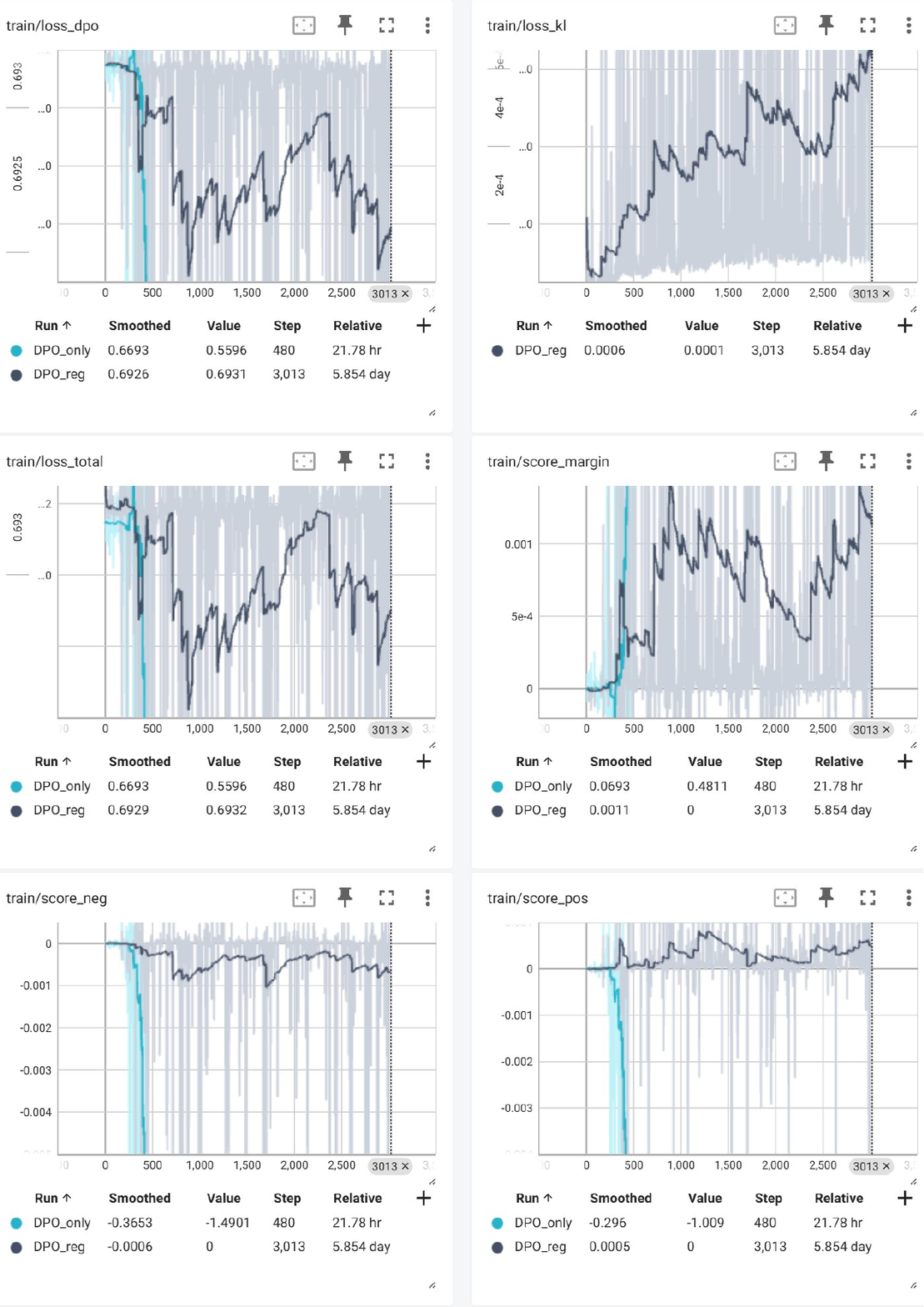}

\caption{\textbf{Training curves} of $\mathcal{L}_\text{DPO} + \lambda_\text{ZO} \mathcal{L}_\text{ZO}$ (black) and $\mathcal{L}_\text{DPO}$ (\textcolor{blue}{blue}).
With only $\mathcal{L}_\text{DPO}$, the training exhibits \textit{likelihood displacement}:
the winner-side implicit reward \textit{score\_pos} drops catastrophically and the loser-side reward \textit{score\_neg} drops even further,
raising up the \textit{score\_margin} in an incorrect way.
In contrast, the existence of $\mathcal{L}_\text{ZO}$ keeps \textit{score\_pos} at slight growth and \textit{score\_neg} at slight drop,
giving a healthier growing of \textit{score\_margin}.}
\label{fig:loss_curve}
\end{figure}
\paragraph{Ablations on the loss setting with training curves.}
This is an extension of~\cref{fig:ablation_loss_visual,sec:ablations} in the main paper.
We show the training curves of DPO loss $\mathcal{L}_\text{DPO}$ and with the regularization losses $\mathcal{L}_\text{ZO}$ in~\cref{fig:loss_curve}.
It can be observed that with only DPO loss, the training is unstable and the loss fluctuates heavily.
With our zeroth-order regularization, the DPO process exhibits more stable training.

\begin{figure}[tb!]
\centering

\begin{subfigure}[t]{0.48\linewidth}
    \centering
    \includegraphics[width=\linewidth]{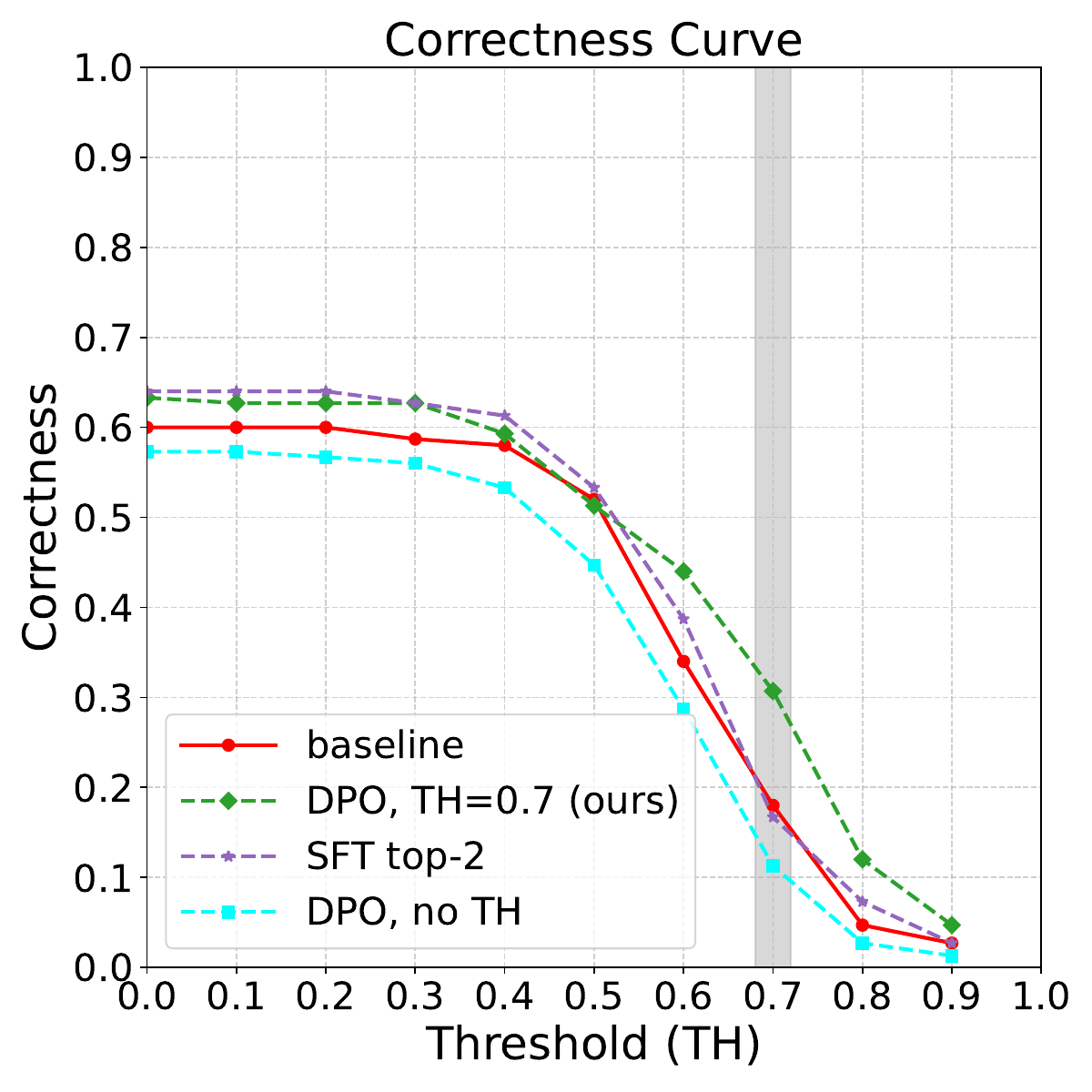}
    \caption{\textbf{Training loss.}}
    \label{fig:ablation_method}
\end{subfigure}
\begin{subfigure}[t]{0.48\linewidth}
    \centering
    \includegraphics[width=\linewidth]{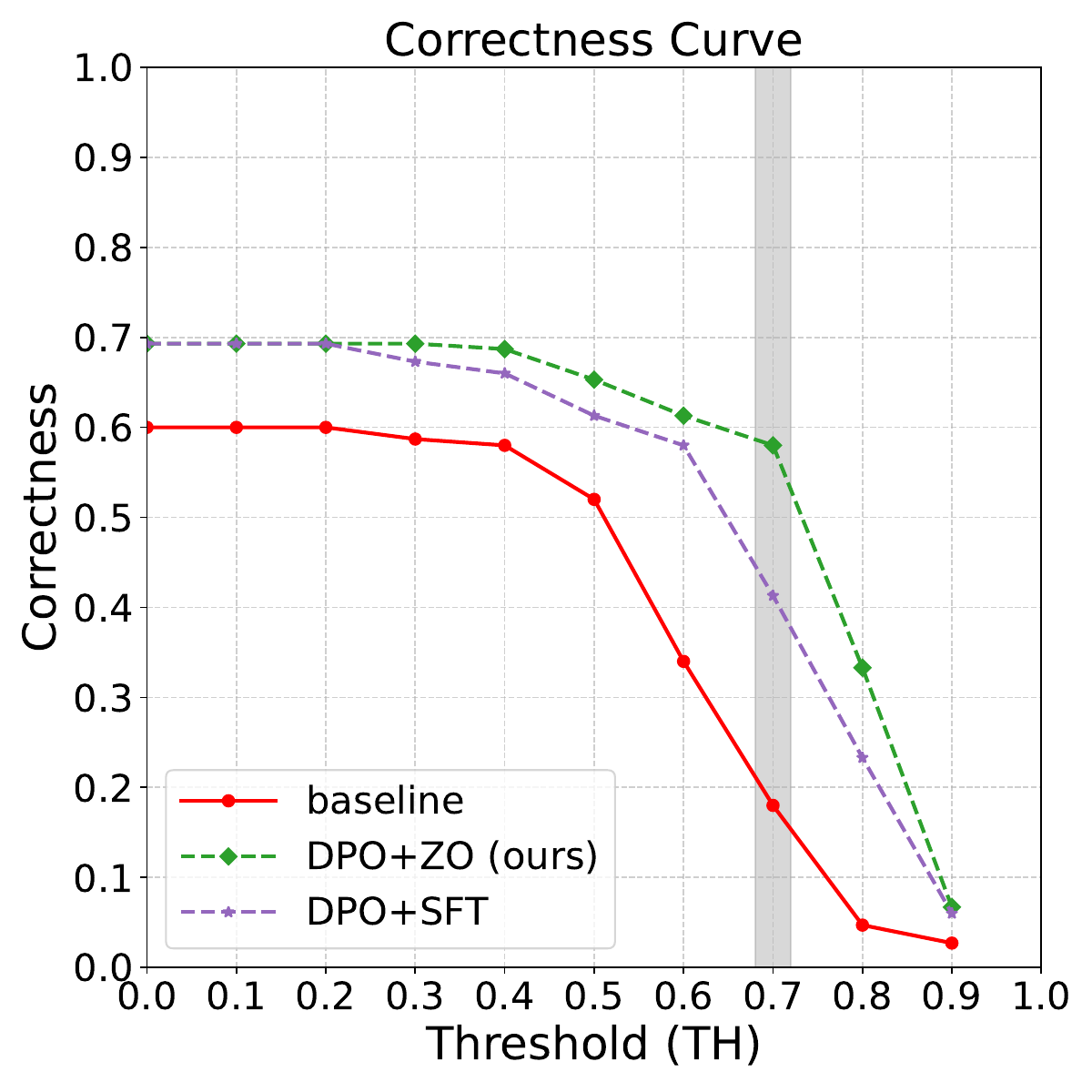}
    \caption{\textbf{Training loss.} (2400 steps)}
    \label{fig:ablation_loss}
\end{subfigure}
\begin{subfigure}[t]{0.48\linewidth}
    \centering
    \includegraphics[width=\linewidth]{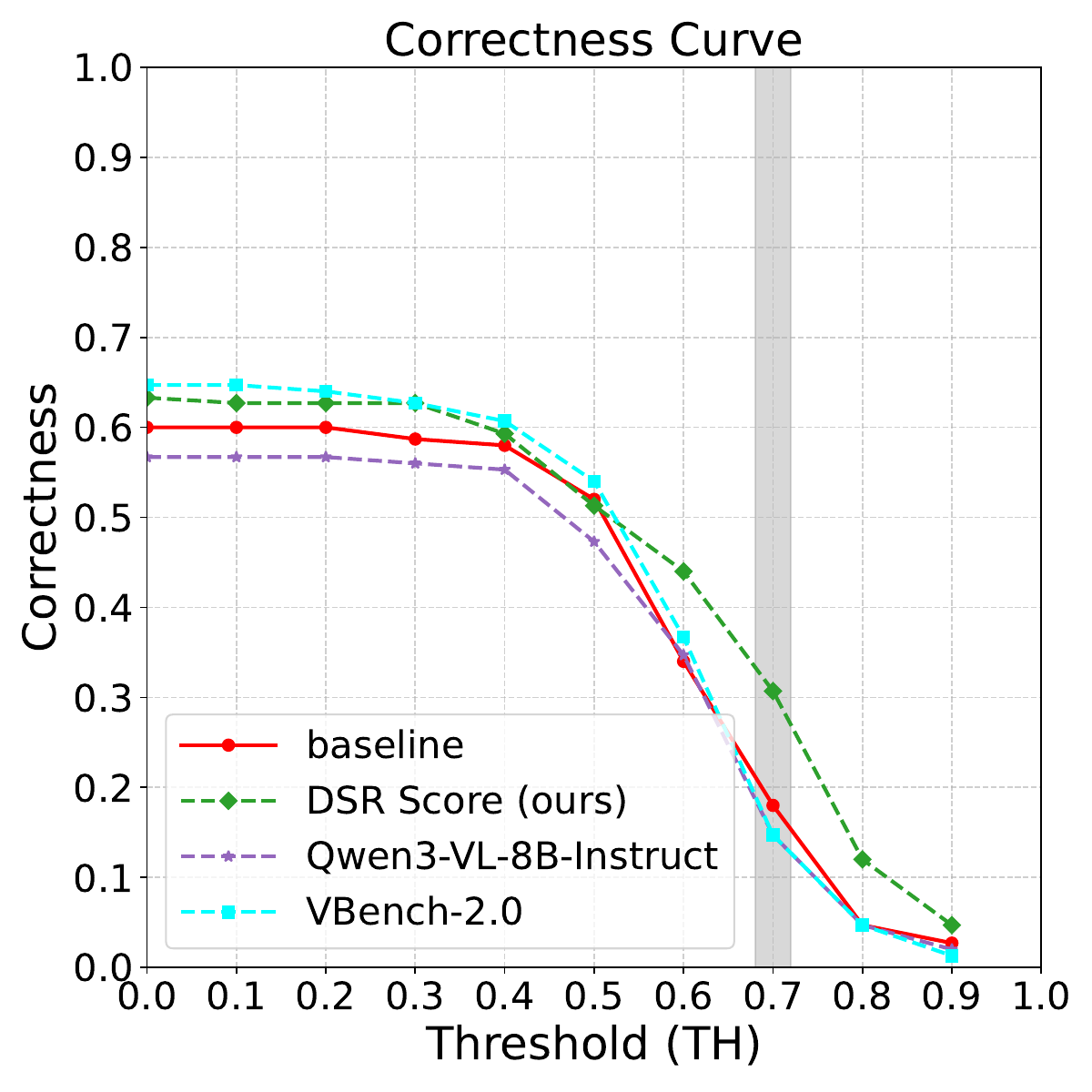}
    \caption{\textbf{Reward System.}}
    \label{fig:ablation_vlmjudge}
\end{subfigure}
\begin{subfigure}[t]{0.48\linewidth}
    \centering
    \includegraphics[width=\linewidth]{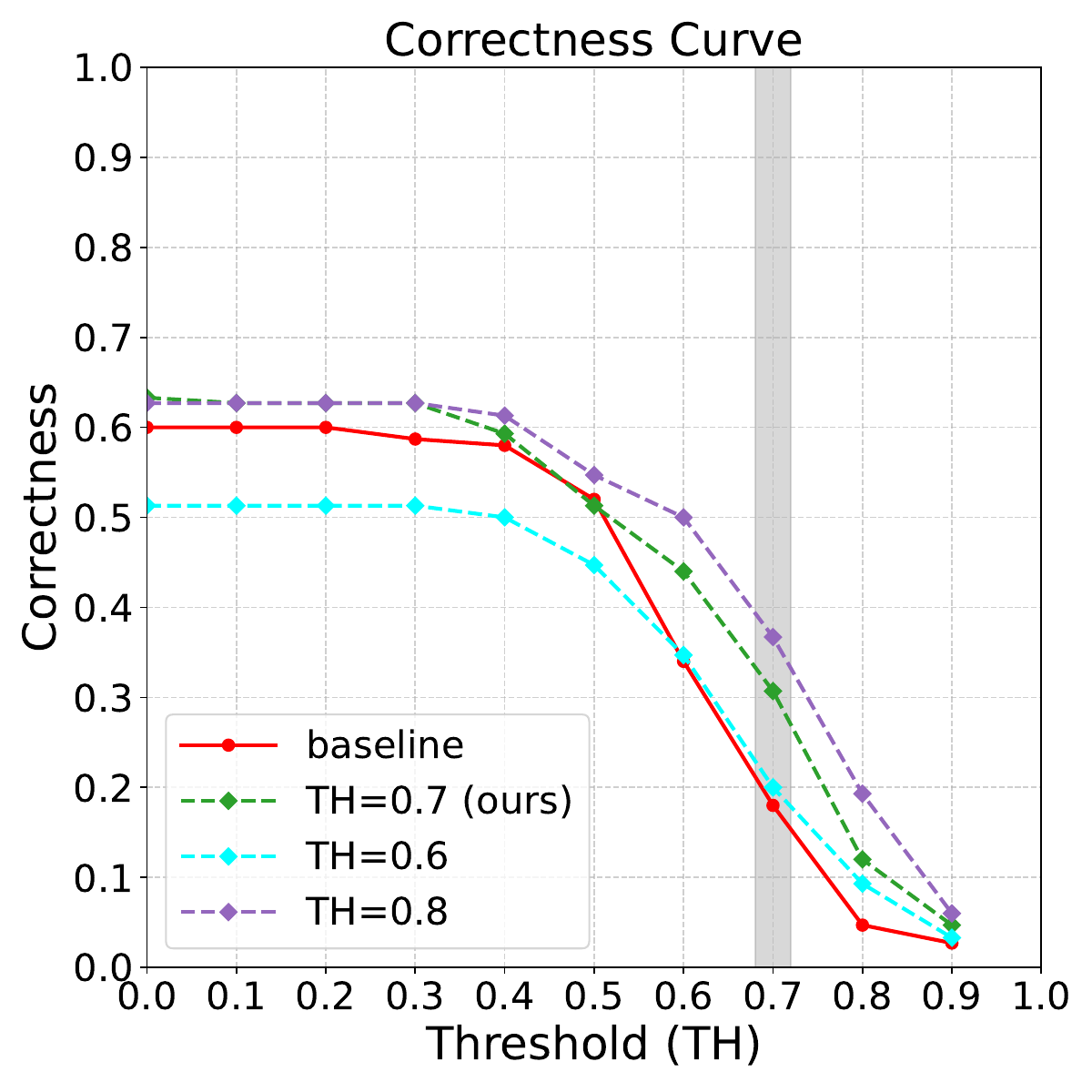}
    \caption{\textbf{Threshold $\tau_{train}$}.}
    \label{fig:ablation_trainTH}
\end{subfigure}
\begin{subfigure}[t]{0.48\linewidth}
    \centering
    \includegraphics[width=\linewidth]{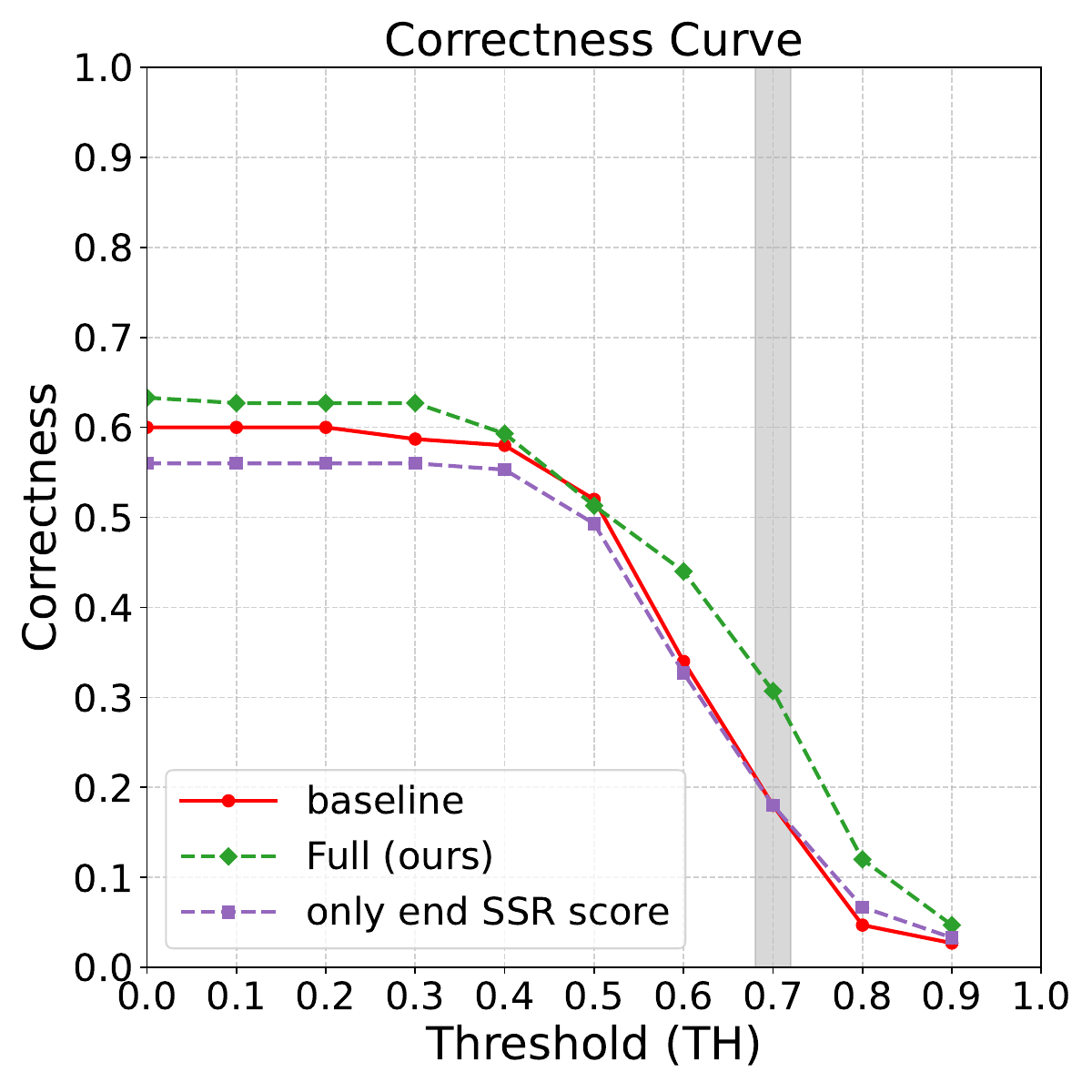}
    \caption{\textbf{Score component.}}
    \label{fig:ablation_dsr_component}
\end{subfigure}
\begin{subfigure}[t]{0.48\linewidth}
    \centering
    \includegraphics[width=\linewidth]{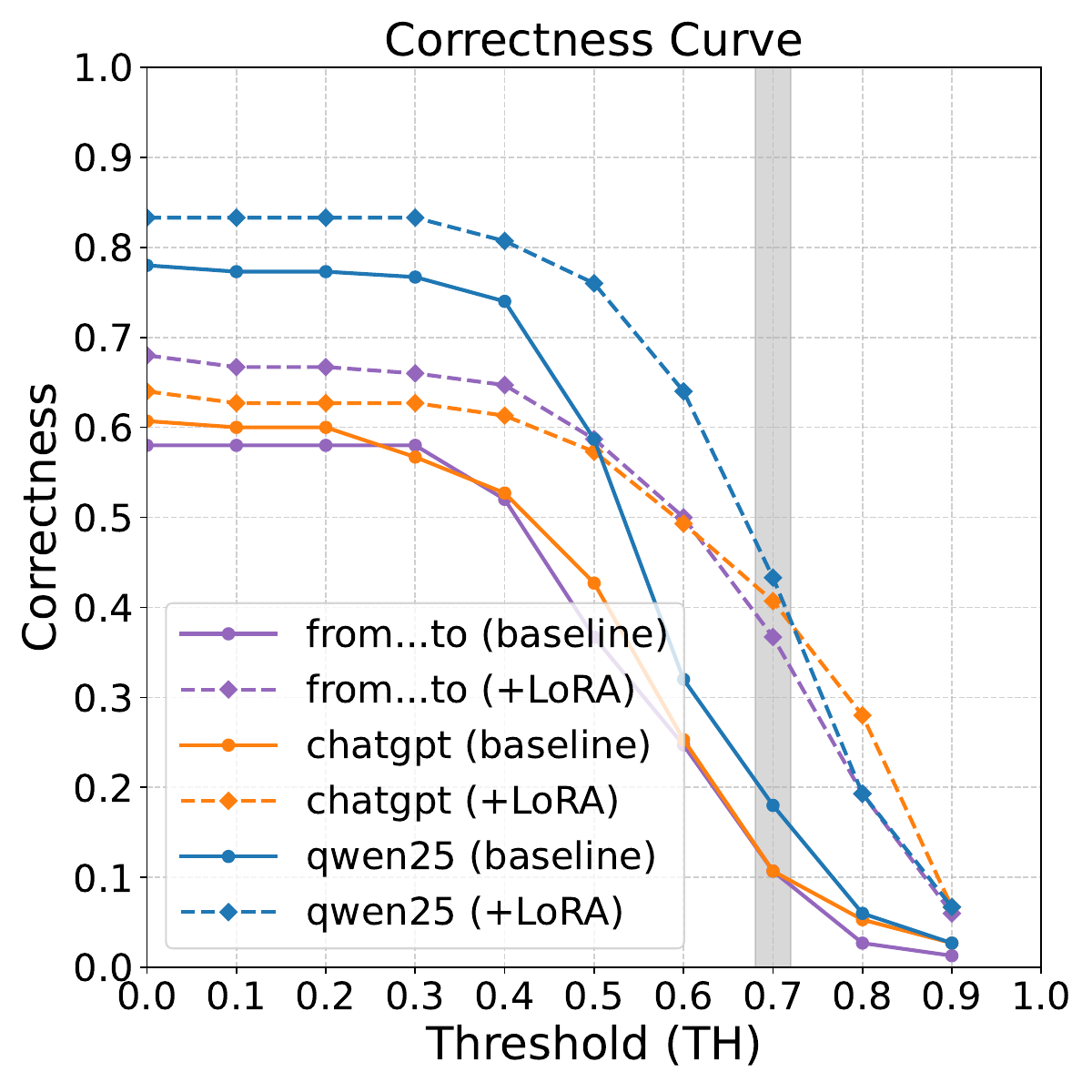}
    \caption{\textbf{Alternative prompt.}}
    \label{fig:ablation_prompt}
\end{subfigure}

\caption{\textbf{Ablation Studies}.
This is an extended version of~\cref{tab:ablation} in the main paper, showing the correctness curves of various ablations.}
\label{fig:ablation_correctness_curve}
\end{figure}
\section{Correctness Curves of Ablation Studies}
\label{sec:ablation_curve}

We show the correctness curves of the ablation studies discussed in~\cref{sec:ablations} to convey additional information beyond the singular values provided in the main paper.

\Cref{fig:ablation_method,fig:ablation_loss} show the ablation under different training losses.
Our setting of doing DPO with $\tau_\text{train}$ shows a clear advantage over SFT/DPO w/o $\tau_\text{train}$.
As for the choice of regularization loss, our $\mathcal{L}_\text{ZO}$ performs better than $\mathcal{L}_\text{SFT}$ as the training progresses to 2400 steps.

\Cref{fig:ablation_trainTH} shows the ablation on different $\tau_\text{train}$ for splitting the winner/loser training samples.
As $\tau_\text{train}$ goes from 0.6 to 0.8, an overall shift is witnessed across all $\tau_\text{test}$.
This shows that our \ourmatric can serve as a stable and progressive optimization signal.

\Cref{fig:ablation_dsr_component,fig:ablation_vlmjudge}
show the ablation using different reward systems.
With only endpoint values, fine-tuning is unable to reach the same level of performance as using the full \ourmatric. Compared to VLM-based evaluation, our \ourmatric gives a much better performance,
suggesting that our metric is more reliable in terms of providing an optimization signal for fine-tuning.

\Cref{fig:ablation_prompt} shows the result of testing on different prompt structures, despite only fine-tuning our model with a specific prompt structure.
For all three types of alternative prompts, the fine-tuned model performs better than the baseline.
This suggests that the fine-tuned model has acquired some semantic knowledge about spatial relationships, rather than simply overfitting to the training prompts.

\section{Theoretical Analysis of the Regularization Term $\mathcal{L}_{\mathrm{ZO}}$}
\label{sec:theory}
\setcounter{equation}{0}

The original DPO loss is
\begin{align}
\mathcal{L}_{\mathrm{DPO}}(\theta)
&= -\mathbb{E}_{(p,x^w,x^l,t)}
\Big[
\log \sigma \Big(
\beta \big( \notag \\
&
(\|\epsilon^w-\epsilon_\theta(x_t^w,t)\|_2^2
-\|\epsilon^w-\epsilon_{\mathrm{ref}}(x_t^w,t)\|_2^2) \notag \\
&
-(\|\epsilon^l-\epsilon_\theta(x_t^l,t)\|_2^2
-\|\epsilon^l-\epsilon_{\mathrm{ref}}(x_t^l,t)\|_2^2)
\big)
\Big)
\Big]
\end{align}

Denote
\begin{align}
A_\theta^{w} &= \|\epsilon^{w}-\epsilon_{\theta}(x_t^{w},t)\|_2^2 -\|\epsilon^{w}-\epsilon_{\mathrm{ref}}(x_t^{w},t)\|_2^2 \\[2pt]
A_\theta^{l} &= \|\epsilon^{l}-\epsilon_{\theta}(x_t^{l},t)\|_2^2 -\|\epsilon^{l}-\epsilon_{\mathrm{ref}}(x_t^{l},t)\|_2^2 \\[2pt]
z &= \beta\big(A_\theta^{w}-A_\theta^{l}\big) \\[2pt]
\mathcal{L}_{\mathrm{DPO}}(\theta) &= -\mathbb{E} \Big[ \log \sigma(z) \Big]
\end{align}
then the gradient of the DPO loss is
\begin{align}
\nabla_\theta \mathcal{L}_{\mathrm{DPO}}(\theta) &= -\mathbb{E}
\Big[
(1-\sigma(z))\,\nabla_\theta z
\Big]
\end{align}

where the gradient $\nabla_\theta z$ is
\begin{align}
\nabla_\theta z &= \beta\big(\nabla_\theta A_\theta^{w} - \nabla_\theta A_\theta^{l} \big) \\[2pt]
\nabla_\theta A_\theta^{w} & = 2\,(\nabla_\theta \epsilon_{\theta}(x_t^{w},t))^{\top}\big(\epsilon_{\theta}(x_t^{w},t)-\epsilon^{w}\big) \\[2pt]
\nabla_\theta A_\theta^{l} & = 2\,(\nabla_\theta \epsilon_{\theta}(x_t^{l},t))^{\top}\big(\epsilon_{\theta}(x_t^{l},t)-\epsilon^{l}\big)
\end{align}

Denote
\begin{align}
J^{w} &= \nabla_\theta \epsilon_{\theta}(x_t^{w},t), \quad
J^{l}  = \nabla_\theta \epsilon_{\theta}(x_t^{l},t) \\[2pt]
\delta^{w} & = \delta_{\theta}(x_t^{w},t) = \epsilon_{\theta}(x_t^{w},t) - \epsilon_{ref}(x_t^{w},t) \\[2pt]
\delta^{l} & = \delta_{\theta}(x_t^{l},t) = \epsilon_{\theta}(x_t^{l},t) - \epsilon_{ref}(x_t^{l},t)
\end{align}

At the neighborhood of $\epsilon_{ref}$
\begin{align}
\mathbb{E}\!\left[\epsilon-\epsilon_{\mathrm{ref}}(x_t,t)\ \right]=0
\end{align}
then
\begin{align}
\delta^{w} & \approx \epsilon_{\theta}(x_t^{w},t) - \epsilon^w, \quad
\delta^{l}   \approx \epsilon_{\theta}(x_t^{l},t) - \epsilon^l
\end{align}
and the gradient can be written as
\begin{align}
\nabla_\theta z
 = \beta\big(\nabla_\theta A_\theta^{w}-\nabla_\theta A_\theta^{l}\big) 
 \approx 2\beta\Big((J^{w})^{\top}\delta^{w}-(J^{l})^{\top}\delta^{l}\Big)
\end{align}

The regularization loss is
\begin{align}
\mathcal{L}_{\mathrm{ZO}}(\theta)
& =
\mathbb{E}_{(p,x^{w},x^{l},t)}
\Big[
\|\epsilon_{\theta}(x_t^{w},t)-\epsilon_{\mathrm{ref}}(x_t^{w},t)\|_2^2 \notag \\
& +
\|\epsilon_{\theta}(x_t^{l},t)-\epsilon_{\mathrm{ref}}(x_t^{l},t)\|_2^2
\Big]
\end{align}
its gradient is
\begin{align}
\nabla_\theta \mathcal{L}_{\mathrm{ZO}}
=
2\,
\mathbb{E}
\Big[
(\nabla_\theta \epsilon_{\theta}(x_t^{w},t))^{\top}
\big(\epsilon_{\theta}(x_t^{w},t)-\epsilon_{\mathrm{ref}}(x_t^{w},t)\big)
+ \notag \\
(\nabla_\theta \epsilon_{\theta}(x_t^{l},t))^{\top}
\big(\epsilon_{\theta}(x_t^{l},t)-\epsilon_{\mathrm{ref}}(x_t^{l},t)\big)
\Big]
\end{align}

With $J^w,J^l,\delta^w,\delta^l$, we have
\begin{align}
\nabla_\theta \mathcal{L}_{\mathrm{ZO}}
=
2\,
\mathbb{E}
\Big[
(J^{w})^{\top}\delta^{w}
+
(J^{l})^{\top}\delta^{l}
\Big]
\end{align}

Denote
\begin{align}
\delta^{S} &= \frac{1}{2}\big(\delta^{w}+\delta^{l}\big), \quad
\delta^{D} = \frac{1}{2}\big(\delta^{w}-\delta^{l}\big), \notag \\
J^{S} &= \frac{1}{2}\big(J^{w}+J^{l}\big), \quad
J^{D} = \frac{1}{2}\big(J^{w}-J^{l}\big)
\end{align}

Then the two gradient can be written as
\begin{align}
\nabla_\theta z
&\approx
2\beta\Big(
(J^{w})^{\top}\delta^{w}
-
(J^{l})^{\top}\delta^{l}
\Big)
\notag \\
&=
4\beta\Big(
(J^{S})^{\top}\delta^{D}
+
(J^{D})^{\top}\delta^{S}
\Big)
\\[10pt]
\nabla_\theta \mathcal{L}_{\mathrm{ZO}}
&=
2\mathbb{E}\Big[
(J^{w})^{\top}\delta^{w}
+
(J^{l})^{\top}\delta^{l}
\Big]
\notag \\
&=
4\mathbb{E}\Big[
(J^{S})^{\top}\delta^{S}
+
(J^{D})^{\top}\delta^{D}
\Big]
\end{align}

Given the same prompt and the same denoising timestep $t$, the two samples $x^w$ and $x^l$ are located within a neighborhood, making $J^w$ close to $J^l$ in terms of norm. This is even consolidated under the low-rank adaption (LoRA) setting of optimization. Therefore, the common component $J^{S} = \frac{1}{2}\big(J^{w}+J^{l}\big)$ should be dominant over $J^{D} = \frac{1}{2}\big(J^{w}-J^{l}\big)$ in terms of norm. \\

In the gradient of DPO $\nabla_\theta z$, $J^s$ amplifies the difference part $\delta^D$, leading to a shift towards $x^w$. However, the common part $\delta^S$ is not amplified at the same level, since $J^D$ is rather small. This undermines the foundation level of alignment to the prompt. In fact, when we want to shift the model distribution towards the preferred samples, an implicit assumption is that the basic level of alignment to both preferred and unwanted samples should be maintained. \\

In the gradient of $\mathcal{L}_{\mathrm{ZO}}$, the common part $\delta^S$ is reserved by the dominant component $J^S$, \textbf{counteracting} the risk of ``losing the common ground'' induced in $\nabla_\theta z$, which helps to maintain the stability at a coefficient $\lambda_{ZO}$. \\

If $\mathcal{L}_{\mathrm{ZO}}$ only constrains the win part $\epsilon^w_{\theta}$, then the gradient will be 
\begin{align}
    (J^{S})^{\top}\delta^{S}
+
(J^{S})^{\top}\delta^{D}
+
(J^{D})^{\top}\delta^{S}
+
(J^{D})^{\top}\delta^{D}
\end{align}
which contains some part of $\nabla_\theta z$ and the alleviation is not as effective.

\section{More Qualitative Results}
(see next page)
\label{sec:more_qualitative_results_appendix}
\begin{figure*}[t]
\centering

\includegraphics[width=\linewidth]{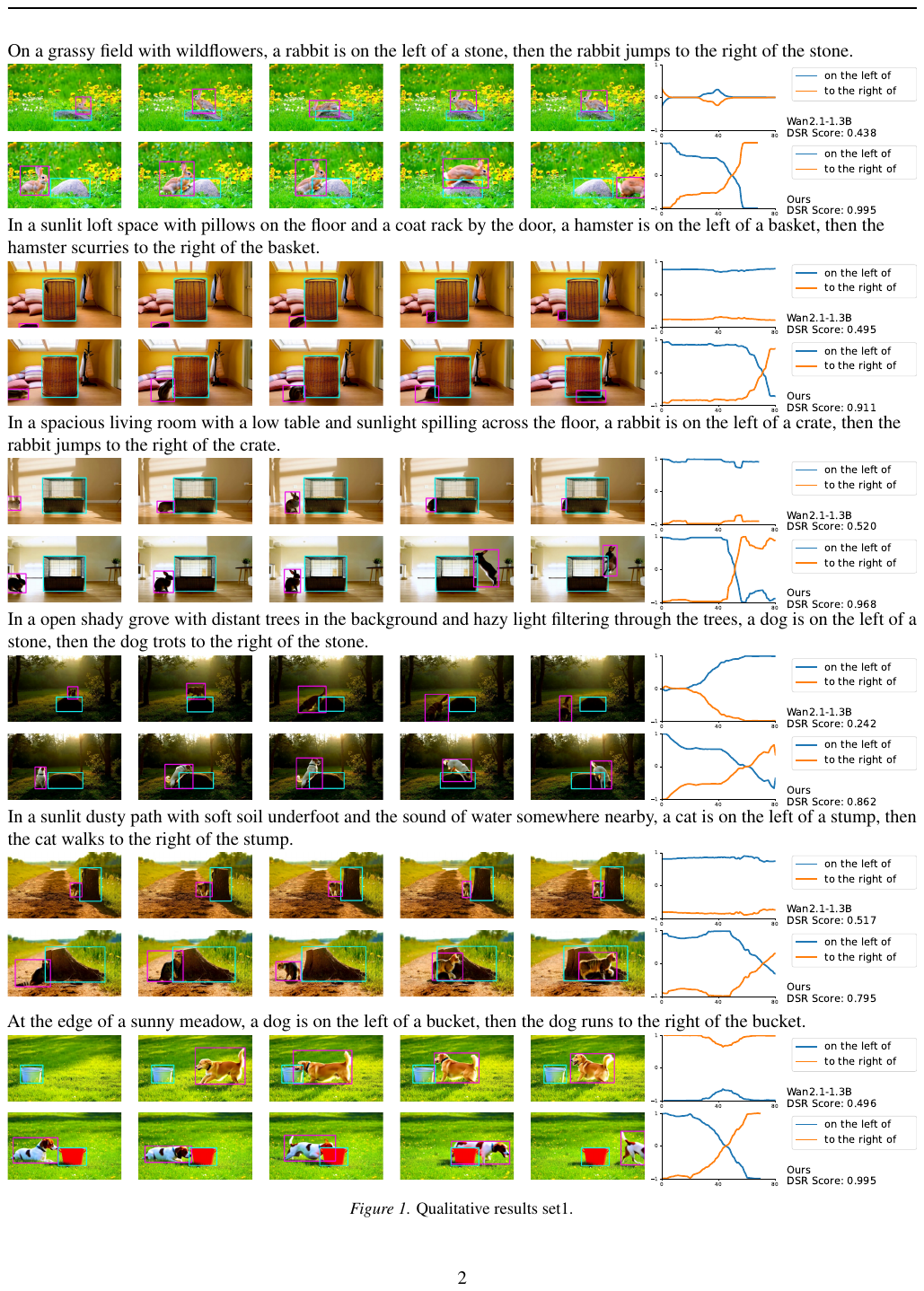}

\caption{Qualitative results set1.}
\label{fig:appendix_qualitative_results_set1}
\end{figure*}

\begin{figure*}[t]
\centering

\includegraphics[width=\linewidth]{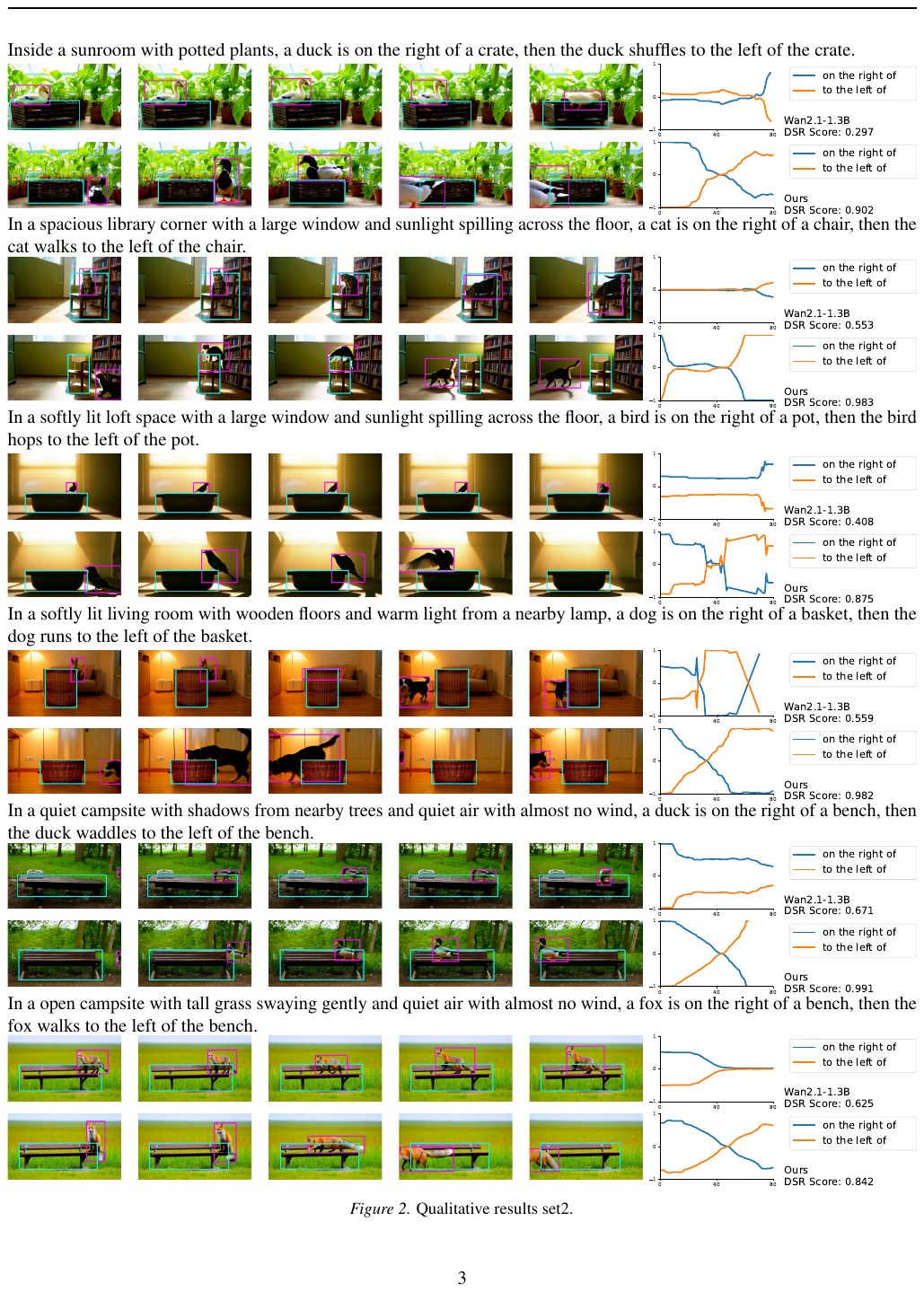}

\caption{Qualitative results set2.}
\label{fig:appendix_qualitative_results_set2}
\end{figure*}

\begin{figure*}[t]
\centering

\includegraphics[width=\linewidth]{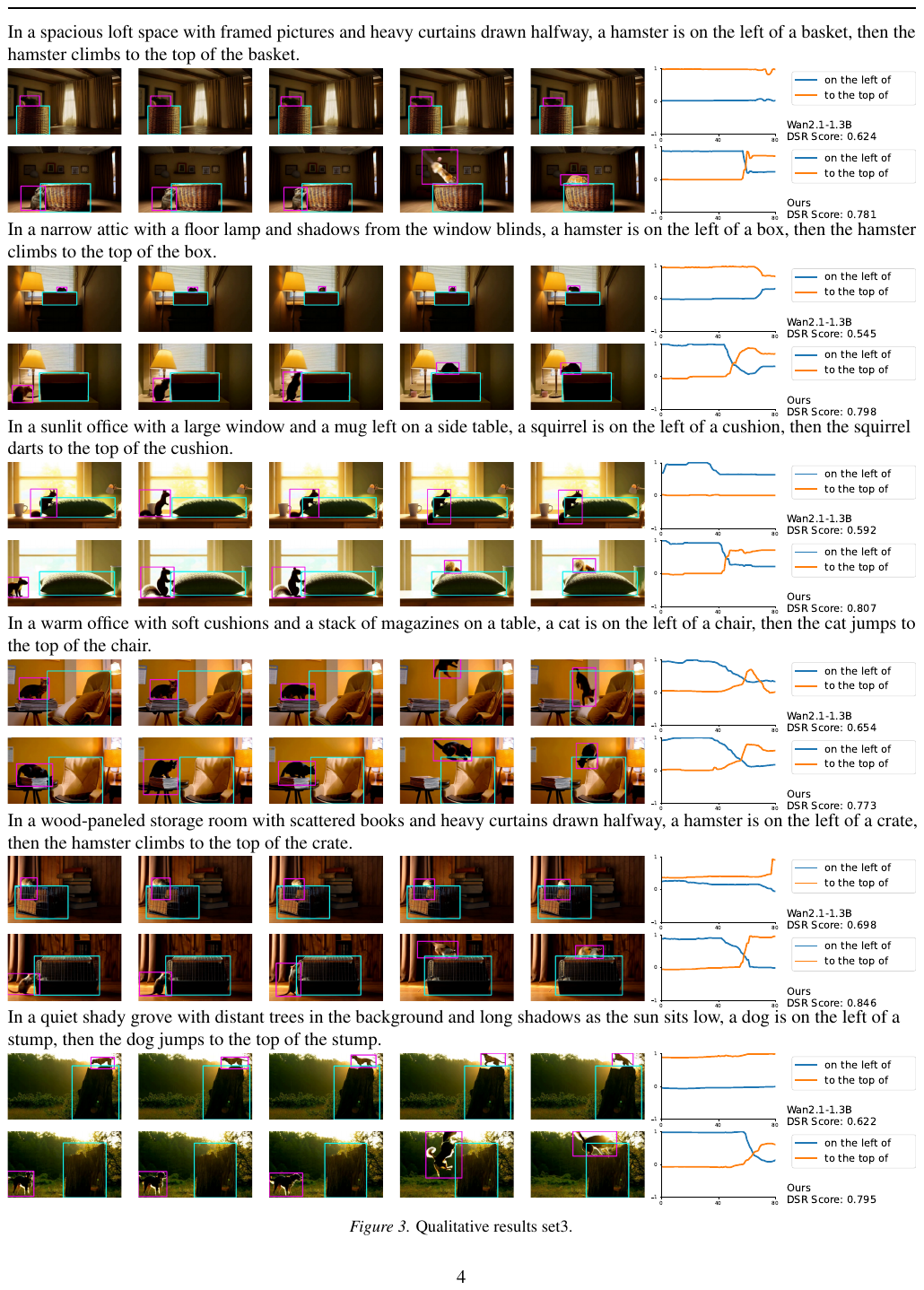}

\caption{Qualitative results set3.}
\label{fig:appendix_qualitative_results_set3}
\end{figure*}

\begin{figure*}[t]
\centering

\includegraphics[width=\linewidth]{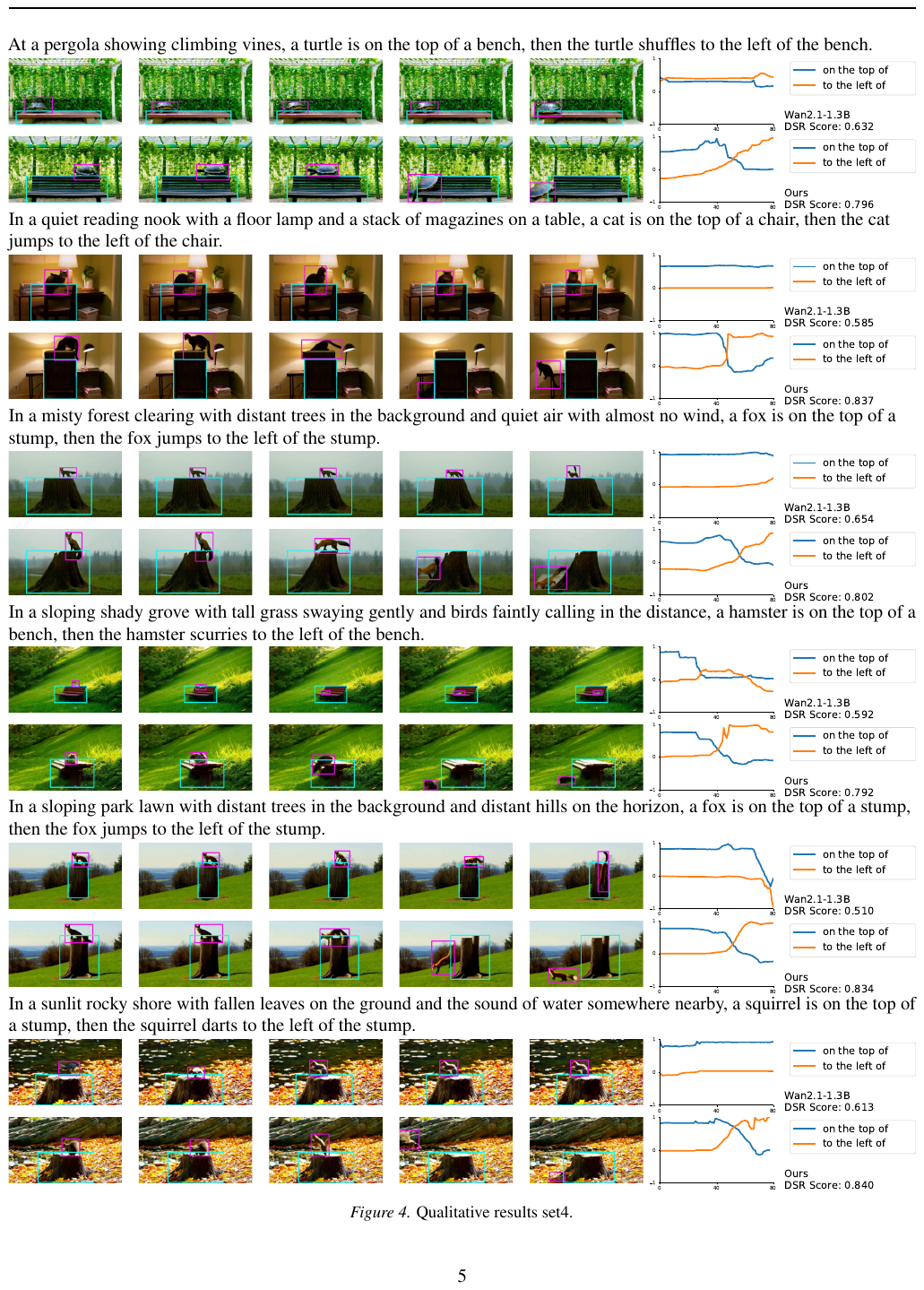}

\caption{Qualitative results set4.}
\label{fig:appendix_qualitative_results_set4}
\end{figure*}

\begin{figure*}[t]
\centering

\includegraphics[width=\linewidth]{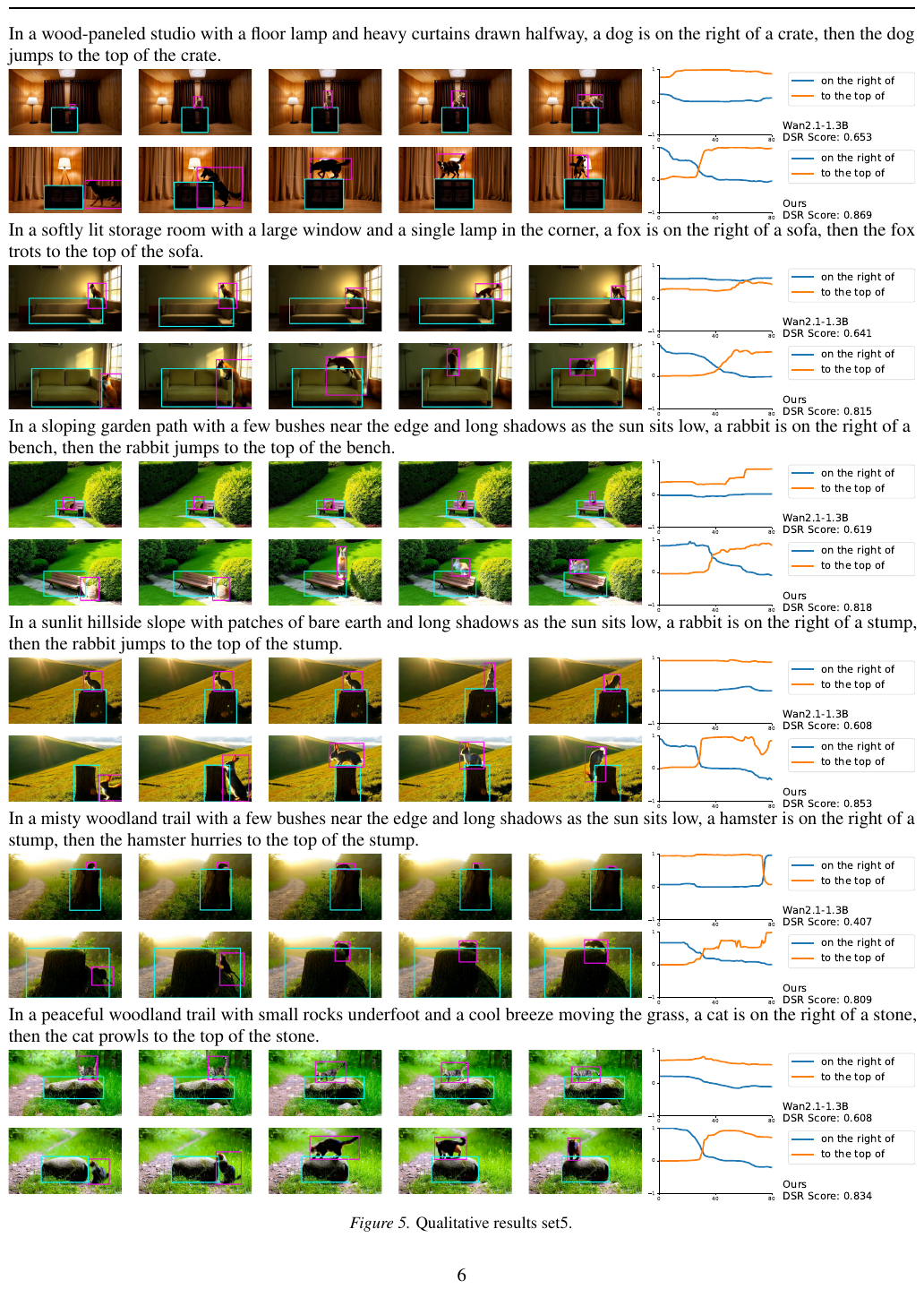}

\caption{Qualitative results set5.}
\label{fig:appendix_qualitative_results_set5}
\end{figure*}

\begin{figure*}[t]
\centering

\includegraphics[width=\linewidth]{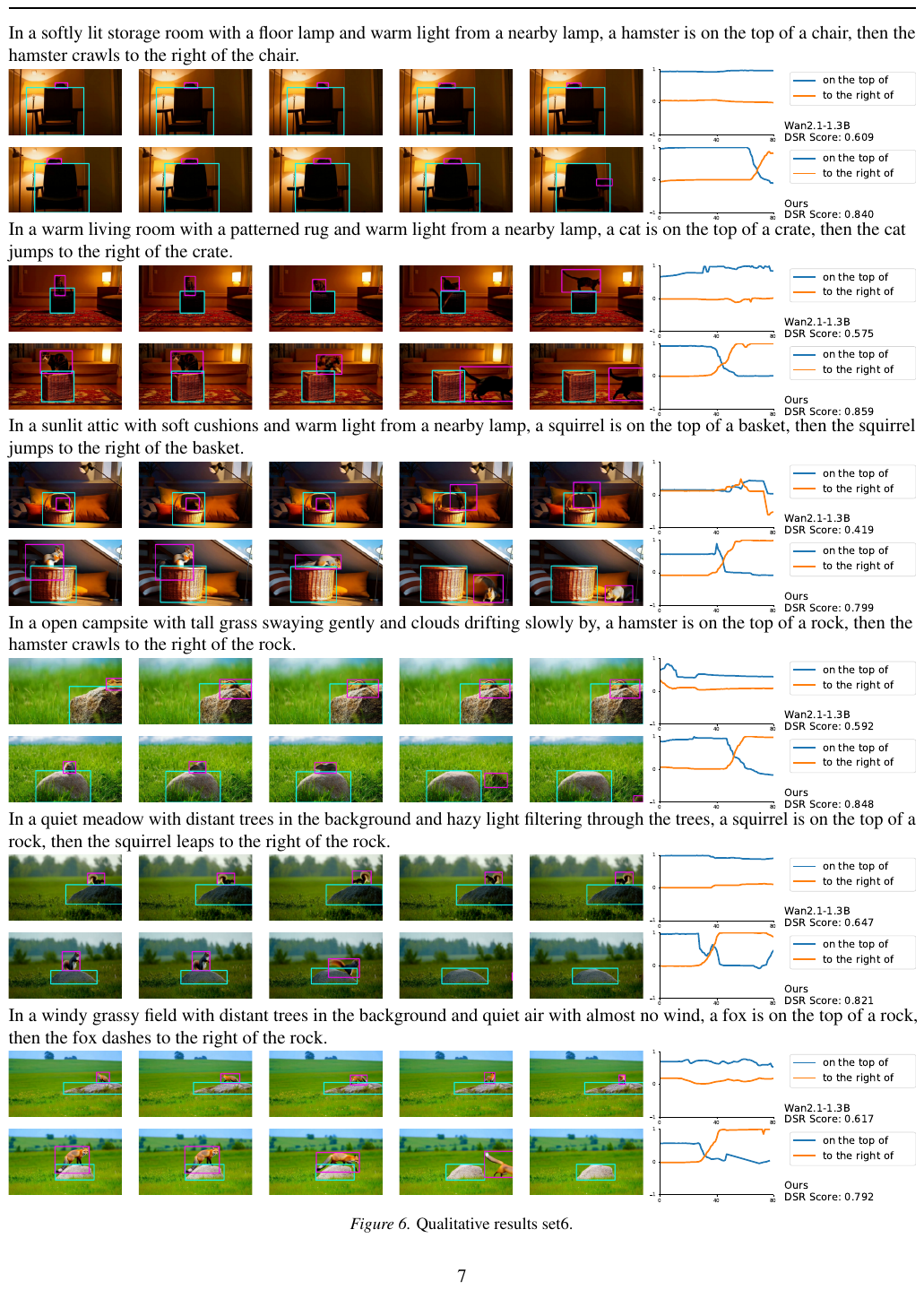}

\caption{Qualitative results set6.}
\label{fig:appendix_qualitative_results_set6}
\end{figure*}

\end{document}